\documentclass{article}

\usepackage{microtype}
\usepackage{graphicx}
\usepackage{subfigure}
\usepackage{booktabs} 

\usepackage{stackengine}

\usepackage{multirow}

\usepackage{amsmath}
\usepackage{amsfonts}
\usepackage{amssymb}
\usepackage{amsthm}

\newcommand{\mb}[1]{\mathbf{#1}}

\newtheorem{theorem}{Theorem}

\usepackage{hyperref}

\usepackage[accepted]{icml2018}

\icmltitlerunning{Attention-based Deep Multiple Instance Learning}

\begin{document}

\twocolumn[
\icmltitle{Attention-based Deep Multiple Instance Learning}



\icmlsetsymbol{equal}{*}

\begin{icmlauthorlist}
\icmlauthor{Maximilian Ilse}{equal,AMLAB}
\icmlauthor{Jakub M.~Tomczak}{equal,AMLAB}
\icmlauthor{Max Welling}{AMLAB}
\end{icmlauthorlist}

\icmlaffiliation{AMLAB}{University of Amsterdam, the Netherlands}

\icmlcorrespondingauthor{Maximilian Ilse}{m.ilse@uva.nl}
\icmlcorrespondingauthor{Jakub M. Tomczak}{j.m.tomczak@uva.nl}

\icmlkeywords{Deep learning, Multiple instance learning, Computational Pathology}

\vskip 0.3in
]



\printAffiliationsAndNotice{\icmlEqualContribution} 

\begin{abstract}
Multiple instance learning (MIL) is a variation of supervised learning where a single class label is assigned to a bag of instances. In this paper, we state the MIL problem as learning the Bernoulli distribution of the bag label where the bag label probability is fully parameterized by neural networks. Furthermore, we propose a neural network-based permutation-invariant aggregation operator that corresponds to the attention mechanism. Notably, an application of the proposed attention-based operator provides insight into the contribution of each instance to the bag label. We show empirically that our approach achieves comparable performance to the best MIL methods on benchmark MIL datasets and it outperforms other methods on a MNIST-based MIL dataset and two real-life histopathology datasets without sacrificing interpretability.
\end{abstract}

\section{Introduction}
\label{introduction}

In typical machine learning problems like image classification it is assumed that an image clearly represents a category (a class). However, in many real-life applications multiple instances are observed and only a general statement of the category is given. This scenario is called \textit{multiple instance learning} (MIL) \cite{DLL:97, ML:98} or, \textit{learning from weakly annotated data} \cite{OBLS:14}. The problem of weakly annotated data is especially apparent in medical imaging \cite{QCCL:17} (\textit{e.g.}, computational pathology, mammography or CT lung screening) where an image is typically described by a single label (benign/malignant) or a Region Of Interest (ROI) is roughly given.

MIL deals with a bag of instances for which a single class label is assigned. Hence, the main goal of MIL is to learn a model that predicts a bag label, \textit{e.g.}, a medical diagnosis. An additional challenge is to discover \textit{key instances} \cite{LIU:12}, \textit{i.e.}, the instances that trigger the bag label. In the medical domain the latter task is of great interest because of legal issues\footnote{According to the European Union General Data Protection Regulation (taking effect 2018), a user should have the right to obtain an explanation of the decision reached.} and its usefulness in clinical practice. In order to solve the primary task of a bag classification different methods are proposed, such as utilizing similarities among bags \cite{CTL:15}, embedding instances to a compact low-dimensional representation that is further fed to a bag-level classifier \cite{ATH:03, CBW:06}, and combining responses of an instance-level classifier \cite{RD:00, RKBDR:08, ZPV:06}. Only the last approach is capable of providing interpretable results. However, it was shown that the instance level accuracy of such methods is low \cite{KH:15} and in general there is a disagreement among MIL methods at the instance level \cite{CSTB:17}. These issues call into question the usability of current MIL models for interpreting the final decision.

In this paper, we propose a new method that aims at incorporating interpretability to the MIL approach and increasing its flexibility. We formulate the MIL model using the Bernoulli distribution for the bag label and train it by optimizing the log-likelihood function. We show that the application of the Fundamental Theorem of Symmetric Functions provides a general procedure for modeling the bag label probability (the bag score function) that consists of three steps: (i) a transformation of instances to a low-dimensional embedding, (ii) a permutation-invariant (symmetric) aggregation function, and (iii) a final transformation to the bag probability. We propose to parameterize all transformations using neural networks (\textit{i.e.}, a combination of convolutional and fully-connected layers), which increases the flexibility of the approach and allows to train the model in an end-to-end manner by optimizing an unconstrained objective function. Last but not least, we propose to replace widely-used permutation-invariant operators such as the maximum operator $\mathrm{max}$ and the mean operator $\mathrm{mean}$ by a trainable weighted average where weights are given by a two-layered neural network. The two-layered neural network corresponds to the attention mechanism \cite{BCB:14, RE:15}. Notably, the attention weights allow us to find key instances, which could be further used to highlight possible ROIs. In the experiments we show that our model is on a par with the best classical MIL methods on common benchmark MIL datasets, and that it outperforms other methods on a MNIST-based MIL problem as well as two real-life histopathology image datasets. Moreover, in the image datasets we provide empirical evidence that our model can indicate key instances.


\section{Methodology}
\label{methodology}

\subsection{Multiple instance learning (MIL)}
\label{mil}

\textbf{Problem formulation} In the classical (binary) supervised learning problem one aims at finding a model that predicts a value of a target variable, $y \in \{0,1\}$, for a given instance, $\mb{x} \in \mathbb{R}^{D}$. In the case of the MIL problem, however, instead of a single instance there is a bag of instances, $X = \{\mb{x}_{1}, \ldots, \mb{x}_{K}\}$, that exhibit neither dependency nor ordering among each other. We assume that $K$ could vary for different bags. There is also a single binary label $Y$ associated with the bag. Furthermore, we assume that individual labels  exist for the instances within a bag, \textit{i.e.}, $y_{1}, \ldots, y_{K}$ and $y_{k} \in \{0,1\}$, for $k=1,\ldots, K$, however, there is no access to those labels and they remain unknown during training. We can re-write the assumptions of the MIL problem in the following form:
\begin{equation}
Y = 
\begin{cases} 
0 ,& \text{iff } \sum_k y_{k} = 0, \\
1 ,&\text{otherwise} .
\end{cases}
\end{equation}
These assumptions imply that a MIL model must be \textbf{permutation-invariant}. Further, the two statements could be re-formulated in a compact form using the maximum operator:
\begin{equation}
Y = \max_{k} \{ y_{k} \} .
\end{equation}

Learning a model that tries to optimize an objective based on the maximum over instance labels would be problematic at least for two reasons. First, all gradient-based learning methods would encounter issues with vanishing gradients. Second, this formulation is suitable only when an instance-level classifier is used.

In order to make the learning problem easier, we propose to train a MIL model by optimizing the log-likelihood function where the bag label is distributed according to the Bernoulli distribution with the parameter $\theta(X) \in [0,1]$, \textit{i.e.}, the probability of $Y=1$ given the bag of instances $X$.

\textbf{MIL approaches} In the MIL setting the bag probability $\theta(X)$ must be permutation-invariant since we assume neither ordering nor dependency of instances within a bag. Therefore, the MIL problem can be considered in terms of a specific form of the Fundamental Theorem of Symmetric Functions with monomials given by the following theorem \cite{ZKRPSS:17}:
\begin{theorem}\label{theorem:universal_mil}
A scoring function for a set of instances $X$, $S(X) \in \mathbb{R}$, is a symmetric function (\textit{i.e.}, permutation-invariant to the elements in $X)$, if and only if it can be decomposed in the following form:
\begin{equation}\label{eq:universal}
S(X) = g\big{(} \sum_{\mb{x} \in X} f(\mb{x}) \big{)} ,
\end{equation}
where $f$ and $g$ are suitable transformations.
\end{theorem}
This theorem provides a general strategy for modeling the bag probability using the decomposition given in (\ref{eq:universal}). A similar decomposition with $\mathrm{max}$ instead of sum is given by the following theorem \cite{QSMG:17}:
\begin{theorem}\label{theorem:approximated_mil}
For any $\varepsilon > 0$, a Hausdorff continuous symmetric function $S(X) \in \mathbb{R}$ can be arbitrarily approximated by a function in the form $g\big{(} \max_{\mb{x} \in X} f(\mb{x}) \big{)}$, where $\mathrm{max}$ is the element-wise vector maximum operator and $f$ and $g$ are continuous functions, that is:
\begin{equation}
|S(X) - g\big{(} \max_{\mb{x} \in X} f(\mb{x}) \big{)} | < \varepsilon .
\end{equation}
\end{theorem}
The difference between Theorems \ref{theorem:universal_mil} and \ref{theorem:approximated_mil} is that the former is a universal decomposition while the latter provides an arbitrary approximation. Nonetheless, they both formulate a general three-step approach for classifying a bag of instances: (i) a transformation of instances using the function $f$, (ii) a combination of transformed instances using a symmetric (permutation-invariant) function $\sigma$, (iii) a transformation of combined instances transformed by $f$ using a function $g$. Finally, the expressiveness of the score function relies on the choice of classes of functions for $f$ and $g$.

In the MIL problem formulation the score function in both theorems is the probability $\theta(X)$ and  the permutation-invariant function $\sigma$ is referred to as the MIL pooling. The choice of functions $f$, $g$ and $\sigma$ determines a specific approach to modeling the label probability. For a given MIL operator there are two main MIL approaches:
\begin{itemize}
\item[(i)] \textit{The instance-level approach}: The transformation $f$ is an instance-level classifier that returns scores for each instance. Then individual scores are aggregated by MIL pooling to obtain $\theta(X)$. The function $g$ is the identity function.
\item[(ii)] \textit{The embedding-level approach}: The function $f$ maps instances to a low-dimensional embedding. MIL pooling is used to obtain a bag representation that is independent of the number of instances in the bag. The bag representation is further processed by a bag-level classifier to provide $\theta(X)$.
\end{itemize}

It is advocated in \citep{WYTPBL:18} that the latter approach is preferable in terms of the bag level classification performance. Since the individual labels are unknown, there is a threat that the instance-level classifier might be trained insufficiently and it introduces additional error to the final prediction. The embedding-level approach determines a joint representation of a bag and therefore it does not introduce additional bias to the bag-level classifier. On the other hand, the instance-level approach provides a score that can be used to find \textit{key instances} \textit{i.e.}, the instances that trigger the bag label. \citet{LIU:12} were able to show that a model that is successfully detecting key instances is more likely to achieve better bag label predictions. We will show how to modify the embedding-level approach to be interpretable by using a new MIL pooling.

\subsection{MIL with Neural Networks}
\label{deep}

In classical MIL problems it is assumed that instances are represented by features that do not require further processing, \textit{i.e.}, $f$ is the identity. However, for some tasks like image or text analysis additional steps of feature extraction are necessary. Additionally, Theorem \ref{theorem:universal_mil} and \ref{theorem:approximated_mil} indicate that for a flexible enough class of functions we can model any permutation-invariant score function. Therefore, we consider a class of transformations that are parameterized by neural networks $f_{\psi}(\cdot )$ with parameters $\psi$ that transform the $k$-th instance into a low-dimensional embedding, $\mb{h}_k = f_{\psi}(\mb{x}_{k})$, where $\mb{h}_k \in \mathcal{H}$ such that $\mathcal{H} = [0,1]$ for the instance-based approach and $\mathcal{H} = \mathbb{R}^{M}$ for the embedding-based approach.

Eventually, the parameter $\theta(X)$ is determined by a transformation $g_{\phi}:\mathcal{H}^{K} \rightarrow [0,1]$. In the instance-based approach the transformation $g_{\phi}$ is simply the identity, while in the embedding-based approach it could be also parameterized by a neural network with parameters $\phi$. The former approach is depicted in Figure \ref{fig:architectures}(a) and the latter in Figure \ref{fig:architectures}(b) in the Appendix.

The idea of parameterizing all transformations using neural networks is very appealing because the whole approach can be arbitrarily flexible and it can be trained end-to-end by backpropagation. The only restriction is that the MIL pooling must be differentiable.

\subsection{MIL pooling}
\label{pooling}

The formulation of the MIL problem requires the MIL pooling $\sigma$ to be permutation-invariant. As shown in Theorem \ref{theorem:universal_mil} and \ref{theorem:approximated_mil}, there are two MIL pooling operators that ensure the score function (\textit{i.e.}, the bag probability) to be a symmetric function, namely, the maximum operator:
\begin{equation}
\forall_{m=1,\ldots, M} : z_{m} = \max_{k=1,\ldots, K} \{ \mb{h}_{km} \} ,
\end{equation}
and the $\mathrm{mean}$ operator:\footnote{Notice that the weight $\frac{1}{K}$ can be seen as a part of the $f$ function.}
\begin{equation}
\mb{z} = \frac{1}{K}\sum_{k=1}^{K} \mb{h}_{k} .
\end{equation}
In fact, other operators could be used such as, the convex maximum operator (\textit{i.e.}, log-sum-exp) \cite{RD:00}, Integrated Segmentation and Recognition \cite{KRL:91}, noisy-or \cite{ML:98} and noisy-and \cite{KBF:16}. These MIL pooling operators could replace $\max$ in Theorem \ref{theorem:approximated_mil} and proofs would follow in a similar manner (see Supplementary in \cite{QSMG:17} for a detailed proof for the maximum operator). All of these operators are differentiable, hence, they could be easily used as a MIL pooling layer in a deep neural network architecture.

\subsection{Attention-based MIL pooling}
\label{attention}

All MIL pooling operators mentioned in the previous section have a clear disadvantage, namely, they are pre-defined and non-trainable. For instance, the $\mathrm{max}$-operator could be a good choice in the instance-based approach but it might be inappropriate for the embedding-based approach. Similarly, the $\mathrm{mean}$ operator is definitely a bad MIL pooling to aggregate instance scores, although, it could succeed in calculating the bag representation. Therefore, a flexible and adaptive MIL pooling could potentially achieve better results by adjusting to a task and data. Ideally, such MIL pooling should also be interpretable, a trait that is missing in all operators mentioned in Section \ref{pooling}.

\textbf{Attention mechanism} We propose to use a weighted average of instances (low-dimensional embeddings) where weights are determined by a neural network. Additionally, the weights must sum to $1$ to be invariant to the size of a bag. The weighted average fulfills the requirements of the Theorem \ref{theorem:universal_mil} where the weights together with the embeddings are part of the $f$ function. Let $H=\{\mb{h}_{1}, \ldots , \mb{h}_{K}\}$ be a bag of $K$ embeddings, then we propose the following MIL pooling:
\begin{equation}\label{eq:weighted_sum}
\mb{z} = \sum_{k=1}^{K} a_k \mb{h}_k,
\end{equation}
where:
\begin{equation}\label{eq:attention}
a_{k} = \dfrac{ \exp\{\mb{w}^{\top} \tanh \big{(} \mb{V} \mb{h}_{k}^{\top} \big{)}\}}{{\displaystyle \sum_{j=1}^K} \exp \{ \mb{w}^{\top} \tanh \big{(}\mb{V} \mb{h}_{j}^{\top}\big{)} \} },
\end{equation}
where $\mb{w} \in \mathbb{R}^{L\times 1}$ and $\mb{V} \in \mathbb{R}^{L \times M}$ are parameters. Moreover, we utilize the hyperbolic tangent $\mathrm{tanh(\cdot )}$ element-wise non-linearity to include both negative and positive values for proper gradient flow. The proposed construction allows to discover (dis)similarities among instances.

Interestingly, the proposed MIL pooling corresponds to a version of the attention mechanism \cite{LFSYXZB:17, RE:15}. The main difference is that typically in the attention mechanism all instances are sequentially dependent while here we assume that all instances are independent. Therefore, a naturally arising question is whether the attention mechanism could work without sequential dependencies among instances, and if it will not learn the $\mathrm{mean}$ operator. We will address this issue in the experiments.

\textbf{Gated attention mechanism} Furthermore, we notice that the $\mathrm{tanh(\cdot)}$ non-linearity could be inefficient to learn complex relations. Our concern follows from the fact that $\tanh(x)$ is approximately linear for  $x \in [-1$, $1$], which could limit the final expressiveness of learned relations among instances. Therefore, we propose to additionally use the gating mechanism \cite{DFAG:16} together with $\mathrm{tanh(\cdot )}$ non-linearity that yields:
\begin{equation}\label{eq:gated_attention}
a_{k} =  \dfrac{\exp \{ \mb{w}^{\top} \big{(} \tanh \big{(} \mb{V} \mb{h}_k^{\top} \big{)} \odot \mathrm{sigm} \big{(} \mb{U} \mb{h}_k^{\top} \big{)} \big{)} \} }{ {\displaystyle \sum_{j=1}^{K}} \exp \{ \mb{w}^{\top} \big{(} \tanh \big{(} \mb{V} \mb{h}_j^{\top} ) \odot \mathrm{sigm} \big{(} \mb{U} \mb{h}_j^{\top} \big{)} \big{)} \} },
\end{equation}
where $\mb{U} \in \mathbb{R}^{L \times M}$ are parameters, $\odot$ is an element-wise multiplication and $\mathrm{sigm(\cdot)}$ is the sigmoid non-linearity. The gating mechanism introduces a learnable non-linearity that potentially removes the troublesome linearity in $\tanh(\cdot)$.

\textbf{Flexibility} In principle, the proposed attention-based MIL pooling allows to assign different weights to instances within a bag and hence the final representation of the bag could be highly informative for the bag-level classifier. In other words, it should be able to find key instances. Moreover, application of the attention-based MIL pooling together with the transformations $f$ and $g$ parameterized by neural networks makes the whole model fully differentiable and adaptive. These two facts make the proposed MIL pooling a potentially very flexible operator that could model an arbitrary permutation-invariant score function. The proposed attention mechanism together with a deep MIL model is depicted in Figure \ref{fig:architectures}(c) in the Appendix.

\textbf{Interpretability} Ideally, in the case of a positive label ($Y=1$), high attention weights should be assigned to instances that are likely to have label $y_k=1$ (key instances). Namely, the attention mechanism allows to easily interpret the provided decision in terms of instance-level labels. In fact, the attention network does not provide scores as the instance-based classifier does but it can be considered as a proxy to that. The attention-based MIL pooling bridges the instance-level approach and the embedding-level approach.

From the practical point of view, \textit{e.g.}, in the computational pathology, it is desirable to provide ROIs together with the final diagnosis to a doctor. Therefore, the attention mechanism is potentially of great interest in practical applications. 

\section{Related work}
\label{relatedWork}

\textbf{MIL pooling} Typically, MIL approaches utilize either the $\mathrm{mean}$ pooling or the $\mathrm{max}$ pooling,  while the latter is mostly used \cite{FZ:17, PC:15, ZLVX:17}. Both operators are non-trainable which potentially limits their applicability. There are MIL pooling operators that contain global adaptive parameters, such as noisy-and \cite{KBF:16}, however, their flexibility is restricted. We propose a fully trainable MIL pooling that adapts to new instances.

\textbf{MIL with neural networks} In the classical work on MIL it is assumed that instances are represented by precomputed features and there is very little need to apply additional feature extraction. Nevertheless, recent work on utilizing fully-connected neural networks in MIL shows that it could still be beneficial \cite{WYTPBL:18}. Similarly, in computer vision the idea of MIL combined with deep learning significantly improves final accuracy \cite{OBLS:14}. In this paper, we follow this line of research since it allows to apply a flexible class of transformations that can be trained end-to-end by backpropagation.

\textbf{MIL and attention} The attention mechanism is widely used in deep learning for image captioning \cite{XBKCCSZB:15} or text analysis \cite{BCB:14, LFSYXZB:17}. In the context of the MIL problem it has rarely been used and only in a very limited form. In \cite{PP:14} an attention-based MIL was proposed but attention weights were trained as parameters of an auxiliary linear regression model. This idea was further expanded and the linear regression model was replaced by a one-layer neural network with single output \cite{PP:17}. The attention-based MIL operator was used very recently in \cite{QSMG:17}, however, the attention was calculated using the dot product and it performed worse than the $\mathrm{max}$ operator. Here, we propose to use a two-layered neural network to learn the MIL operator and we show that it outperforms commonly used MIL pooling operators.

\textbf{MIL for medical imaging} The MIL seems to perfectly fit medical imaging where processing a whole image consisting of billions of pixels is computationally infeasible. Moreover, in the medical domain it is very difficult to obtain pixel-level annotations, that drastically reduces number of available data. Therefore, it is tempting to divide a medical image into smaller patches that could be further considered as a bag with a single label \cite{QCCL:17}. This idea attracts a great interest in the computational histopathology where patches could correspond to cells that are believed to indicate malignant changes \cite{SRTSCR:16}. Different MIL approaches were used for histopathology data, such as, Gaussian processes \cite{KZH:14, KHDRLH:16} or a two-stage approach with neural networks and EM algorithm to determine instance classes \cite{HSKGDS:16}. Other applications of MIL methods in medical imaging are mammography (nodule) classification \cite{ZLVX:17} and microscopy cell detection \cite{KBF:16}. In this paper, we show that the proposed attention-based deep MIL approach can be used not only to provide the final diagnosis but also to indicate ROIs in a histopathology slide.

\section{Experiments}
\label{experiments}

In the experiments we aim at evaluating the proposed approach: a MIL model parameterized with neural networks and a (gated) attention-based pooling layer ('Attention' and 'Gated-Attention'). We evaluate our approach on a number of different MIL datasets: five MIL benchmark datasets (\textsc{Musk1}, \textsc{Musk2}, \textsc{Fox}, \textsc{Tiger}, \textsc{Elephant}), an MNIST-based image dataset (\textsc{MNIST-bags}) and two real-life histopathology datasets (\textsc{Breast Cancer}, \textsc{Colon Cancer}). We want to verify two research questions in the experiments: (i) whether our approach achieves the best performance or is comparable to the best performing method, (ii) if our method can provide interpretable results by using the attention weights that indicate key instances or ROIs.

In order to obtain a fair comparison we use a common evaluation methodology, \textit{i.e.}, 10-fold-cross-validation, and five repetitions per experiment. In the case of \textsc{MNIST-bags} we use a fixed division into training and test set. In order to create test bags we solely sampled images from the MNIST test set. During training we only used images from the MNIST training set. For all experiments we use modified versions of models that have shown high classification performance on the individual datasets \cite{WYTPBL:18,LBBH:98, SRTSCR:16}. The MIL pooling layers are either located before the last layer of the model (the embedded-based approach) or after last layer of the model (the instance-based approach). If an attention-based MIL pooling layer is used the number of parameters in \textbf{V} was determined using a validation set. We tested the following dimensions ($L$): 64, 128 and 256. The different dimensions only resulted in minor changes of the model's performance. For layers using the gated attention mechanism \textbf{V} and \textbf{U} have the same number of parameters. Finally, all layers were initialized according to \citet{GB:10} and biases were set to zero.

We compare our approach to various MIL methods on MIL benchmark datasets. On the image datasets our method is compared with instance-level and embedding-level neural networks and commonly used MIL pooling layers ($\mathrm{max}$ and $\mathrm{mean}$). In the following, we are using 'Instance+max/mean' and 'Embedding+max/mean' to indicate networks that are build from convolutional layers and fully-connected layers. In contrast to networks purely build from fully-connected layers, referred to as 'mi-Net' and 'MI-Net' \cite{WYTPBL:18}.

On \textsc{MNIST-bags} we include a SVM-based MIL model, called (\textsc{MI-SVM}). We do not present results of \textsc{MI-SVM} on the histopathology datasets since we could not train (including hyperparameter search and five times 10-fold-cross-validation procedure) the model in a reasonable amount of time.\footnote{Learning a single MI-SVM took approximately one week due to the large number of patches.} In order to compare the bag level performance we use the following metrics: the classification accuracy, precision, recall, F-score, and the area under the receiver operating characteristic curve (AUC).

\subsection{Classical MIL datasets}
\label{classical}
\textbf{Details} In the first experiment we aim at verifying whether our approach can compete with the best MIL methods on historically important benchmark datasets. Since all five datasets contain precomputed features and only a small number of instances and bags, neural networks are most likely not well suited. First we predict drug activity (\textsc{Musk1} and \textsc{Musk2}). A molecule has the desired drug effect if and only if one or more of its conformations bind to the target binding site. Since molecules can adopt multiple shapes, a bag is made up of shapes belonging to the same molecule \cite{DLL:97}. The three remaining datasets, \textsc{Elephant}, \textsc{Fox} and \textsc{Tiger}, contain features extracted from images. Each bag consists of a set of segments of an image. For each category, positive bags are images that contain the animal of interest, and negative bags are images that contain other animals \cite{ATH:03}. For detailed information on the number of bags, instances and features in each dataset see Section \ref{app:classic} in the Appendix.

In our experiments we use the same architecture, optimizer and hyperparameters as in the MI-Net model \cite{WYTPBL:18}.
\begin{table}[!htbp]
  \centering
  \caption{Results on classical MIL datasets. Experiments were run 5 times and an average of the classification accuracy ($\pm$ a standard error of a mean) is reported. [1] \cite{ATH:03}, [2] \cite{GFKS:02}, [3] \cite{ZG:02} [4] \cite{ZSL:09} [5] \cite{WWZ:17} [6] \cite{WYTPBL:18}}
  \vspace{2mm}
  \resizebox{1.0\columnwidth}{!}{
  \begin{tabular}{lccccc}
    \toprule
    \textsc{Method}						& \textsc{Musk1} & \textsc{Musk2} & \textsc{Fox} & \textsc{Tiger} & \textsc{Elephant} \\
    \midrule   
    mi-SVM [1]				& 0.874$\pm$N/A\hspace{5pt}	& 0.836$\pm$N/A\hspace{5pt}	& 0.582$\pm$N/A\hspace{5pt}	& 0.784$\pm$N/A\hspace{5pt}	& 0.822$\pm$N/A\hspace{5pt}	 \\ 
    MI-SVM	[1]				& 0.779$\pm$N/A\hspace{5pt}	& 0.843$\pm$N/A\hspace{5pt}	& 0.578$\pm$N/A\hspace{5pt}	& 0.840$\pm$N/A\hspace{5pt}	& 0.843$\pm$N/A\hspace{5pt}	 \\ 
    MI-Kernel [2]			& \textbf{0.880}$\pm$0.031	& \textbf{0.893}$\pm$0.015	& \textbf{0.603}$\pm$0.028	& 0.842$\pm$0.010			& 0.843$\pm$0.016 \\ 
    EM-DD	[3]				& 0.849$\pm$0.044			& \textbf{0.869}$\pm$0.048	& \textbf{0.609}$\pm$0.045	& 0.730$\pm$0.043			& 0.771$\pm$0.043 \\ 
    mi-Graph [4]			& \textbf{0.889}$\pm$0.033	& \textbf{0.903}$\pm$0.039	& \textbf{0.620}$\pm$0.044	& \textbf{0.860}$\pm$0.037	& \textbf{0.869}$\pm$0.035 \\ 
	miVLAD	[5]				& \textbf{0.871}$\pm$0.043	& \textbf{0.872}$\pm$0.042	& \textbf{0.620}$\pm$0.044	& 0.811$\pm$0.039			& \textbf{0.850}$\pm$0.036 \\ 
    miFV	[5]				& \textbf{0.909}$\pm$0.040	& \textbf{0.884}$\pm$0.042	& \textbf{0.621}$\pm$0.049	& 0.813$\pm$0.037			& \textbf{0.852}$\pm$0.036 \\ 
    \midrule  
    mi-Net	[6]				& \textbf{0.889}$\pm$0.039	& \textbf{0.858}$\pm$0.049	& \textbf{0.613}$\pm$0.035  & 0.824$\pm$0.034			& \textbf{0.858}$\pm$0.037 \\ 
    MI-Net	[6]				& \textbf{0.887}$\pm$0.041	& \textbf{0.859}$\pm$0.046	& \textbf{0.622}$\pm$0.038  &  \textbf{0.830}$\pm$0.032	& \textbf{0.862}$\pm$0.034 \\ 
MI-Net with DS [6]			& \textbf{0.894}$\pm$0.042	& \textbf{0.874}$\pm$0.043	& \textbf{0.630}$\pm$0.037	&  \textbf{0.845}$\pm$0.039	& \textbf{0.872}$\pm$0.032 \\ 
MI-Net with RC [6]			& \textbf{0.898}$\pm$0.043	& \textbf{0.873}$\pm$0.044	& \textbf{0.619}$\pm$0.047	&  \textbf{0.836}$\pm$0.037	& \textbf{0.857}$\pm$0.040 \\
    \midrule  
Attention			& \textbf{0.892}$\pm$0.040	& \textbf{0.858}$\pm$0.048	& \textbf{0.615}$\pm$0.043	&  \textbf{0.839}$\pm$0.022 & \textbf{0.868}$\pm$0.022 \\ 
Gated-Attention	& \textbf{0.900}$\pm$0.050	& \textbf{0.863}$\pm$0.042	& \textbf{0.603}$\pm$0.029	&  \textbf{0.845}$\pm$0.018 & \textbf{0.857}$\pm$0.027 \\ 
    \bottomrule
  \end{tabular}
  }
  \label{tab:classic_MIL_results}
\end{table}

\textbf{Results and discussion} The results of the experiment are presented in Table \ref{tab:classic_MIL_results}. Our approaches (Attention and Gated-Attention) are comparable with the best performing classical MIL methods (notice the standard error of the mean).

\subsection{MNIST-bags}
\label{mnist}

\textbf{Details} The main disadvantage of the classical MIL benchmark datasets is that instances are represented by precomputed features. In order to consider a more challenging scenario, we propose to investigate a dataset that is created using the well-known MNIST image dataset. A bag is made up of a random number of $28\times28$ grayscale images taken from the MNIST dataset. The number of images in a bag is Gaussian-distributed and the closest integer value is taken. A bag is given a positive label if it contains one or more images with the label '9'. We chose '9' since it can be easily mistaken with '7' or '4'. We investigate the influence of the number of bags in the training set as well as the average number of instances per bag on the prediction performance. During evaluation we use a fixed number of $1000$ test bags. For all experiments a LeNet5 model is used \cite{LBBH:98}, see Table \ref{tab:mnist_bag_embedding_model} and \ref{tab:mnist_bag_instance_model} in the Appendix. The models are trained with the Adam optimization algorithm \cite{KB:14}. We keep the default parameters for $\beta_1$ and $\beta_2$, see Table \ref{tab:mnist_bag_optimizer} in the Appendix. In addition, we compare our method with a SVM-based MIL method (MI-SVM) \cite{ATH:03} that uses a Gaussian kernel on raw pixel features\footnote{We use code provided with \cite{DR:14}: \url{https://github.com/garydoranjr/misvm}}.

In the experiments we use different numbers of the mean bag size, namely, $10$, $50$ and $100$, and the variance $2, 10, 20$, respectively. Moreover, we use varying numbers of training bags, \textit{i.e.}, $50, 100, 150, 200, 300, 400, 500$. These different settings allow us to verify how different number of training bags and different number of instances influence MIL models. We compare instance-based and embedding-based approaches parameterized with a neural network (LeNet5) with $\mathrm{mean}$ and $\mathrm{max}$ MIL pooling. We use AUC as the evaluation metric.
\begin{figure}[!htbp]
	\centering
	\includegraphics[width=0.825\linewidth]{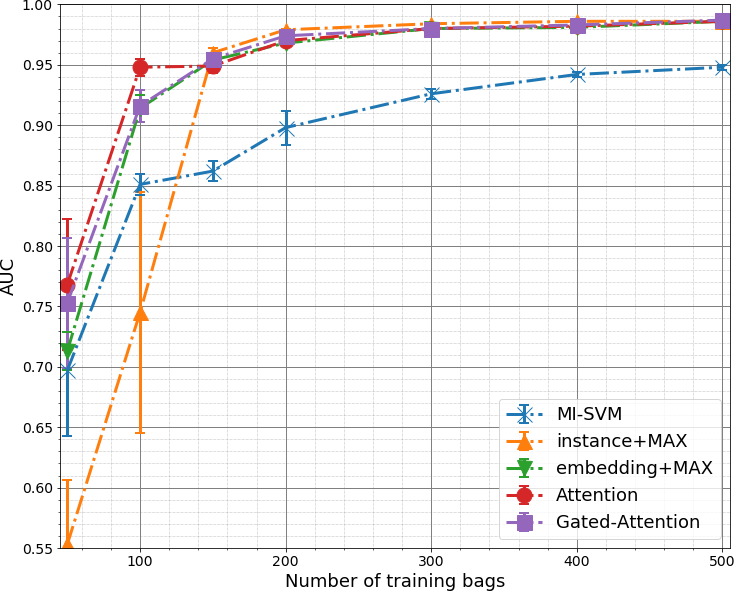}
    \vskip -2mm
	\caption{The test AUC for \textsc{MNIST-bags} with on average $10$ instances per bag.}
	\label{fig:mnist_bags_10}
\end{figure}

\begin{figure}[!htbp]
	\centering
	\includegraphics[width=0.825\linewidth]{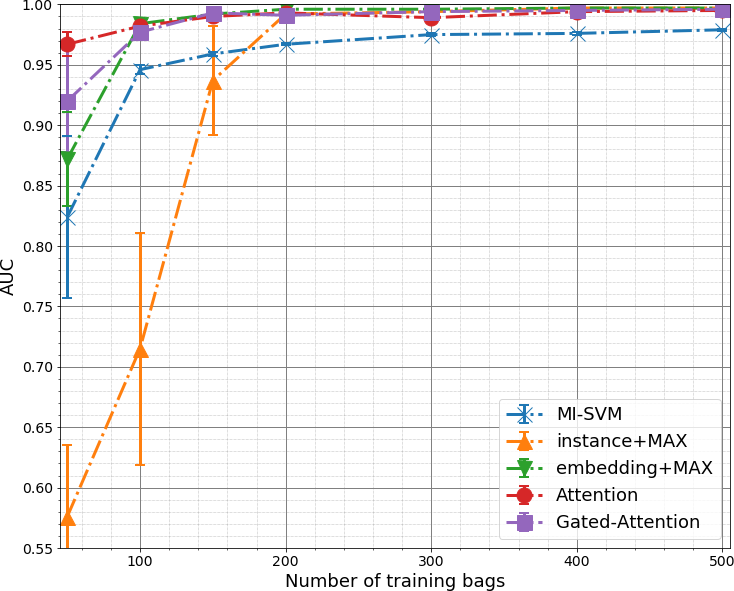}
    \vskip -2mm
	\caption{The test AUC for \textsc{MNIST-bags} with on average $50$ instances per bag.}
	\label{fig:mnist_bags_50}
\end{figure}

\begin{figure}[!htbp]
	\centering
	\includegraphics[width=0.825\linewidth]{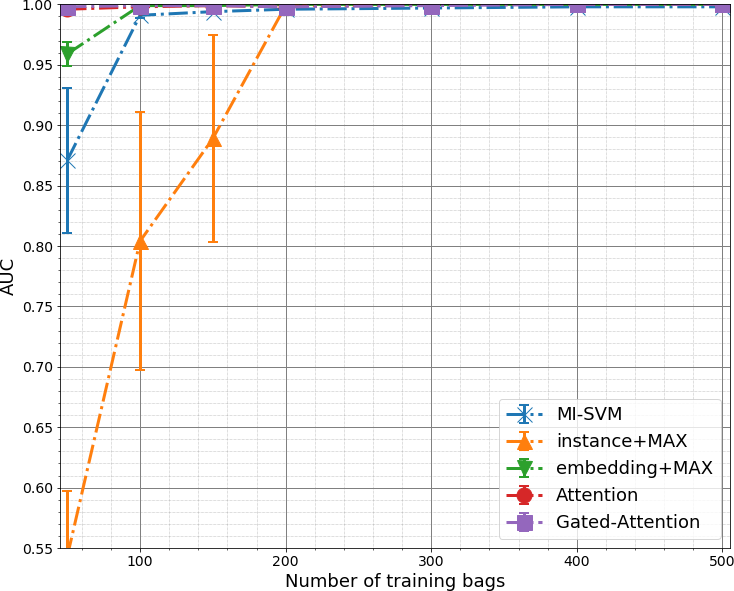}
    \vskip -2mm
	\caption{The test AUC for \textsc{MNIST-bags} with on average $100$ instances per bag.}
	\label{fig:mnist_bags_100}
\end{figure}
\textbf{Results and discussion} The results of AUC for the mean bag sizes equal to $10$, $50$ and $100$ are presented in Figure \ref{fig:mnist_bags_10}, \ref{fig:mnist_bags_50} and \ref{fig:mnist_bags_100}, respectively, and detailed results are given in the Appendix. The findings of the experiment are the following: First, the proposed attention-based deep MIL approach performs much better than other methods in the small sample size regime. Moreover, when there is a small effective size of the training set that corresponds to 50-150 bags for around 10 instances per bag (see Figure \ref{fig:mnist_bags_10}) or 50-100 bags in the case of on average 50 instances in a bag (see Figure \ref{fig:mnist_bags_50}), our method still achieves significantly higher AUC than all other methods. Second, we notice that our approach is more flexible and obtained better results than the SVM-based approach in all cases except large effective sample sizes (see Figure \ref{fig:mnist_bags_100}). Third, the embedding-based models performed better than the instance-based models. However, for a sufficient number of training images (number of training bags and training instances per bag) all models achieve very similar results. Fourth, the $\mathrm{mean}$ operator performs significantly worse than the $\mathrm{max}$ operator. However, the embedding-based model with the $\mathrm{mean}$ operator converged eventually to the best value but always later than the one with $\mathrm{max}$. See Section \ref{app:mnist} in the Appendix for details.

The results of this experiment indicate that for a small-sample size regime our approach is preferable to others. Since attention serves as a gradient update filter during backpropagation \cite{WJQYLZWT:17}, instances with higher weights will contribute more to learning the encoder network of instances. This is especially important since medical imaging problems contain only a small number of cases. In general, the more instances are in a bag the easier the MIL task becomes, since the MIL assumption states that every instance in a negative bag is negative. For example, a negative bag of size 100 from the MNIST-bags dataset will include about 11 negative examples per class.

\begin{figure}[h]
\centering
\includegraphics[width=1.0\linewidth]{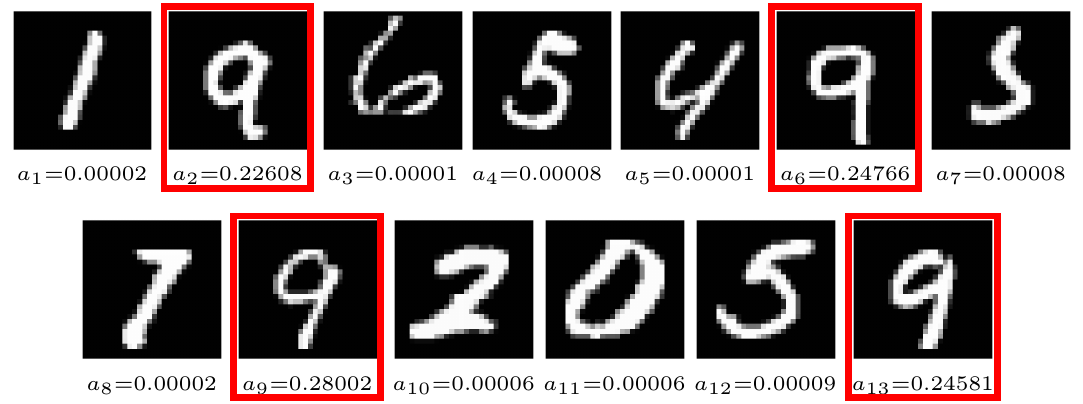}
\caption{Example of attention weights for a positive bag.}
\label{fig:attention_mnist_bags}
\end{figure}

Finally, we present an exemplary result of the attention mechanism in Figure \ref{fig:attention_mnist_bags}. In this example a bag consists of $13$ images. For each digit the corresponding attention weight is given by the trained network. The bag is properly predicted as positive and all nines are correctly highlighted. Hence, the attention mechanism works as expected. More examples are given in the Appendix.

\subsection{Histopathology datasets}
\label{breast}

\textbf{Details} An automatic detection of cancerous regions in hematoxylin and eosin (H$\&$E) stained whole-slide images is a task with high clinical relevance. Current supervised approaches utilize pixel-level annotations \cite{LetAl:17}. However, data preparation requires large amount of time from pathologists which highly interferes with their daily routines. Hence, a successful solution working with weak labels would hold a great promise to reduce the workload of the pathologists. In the following, we perform two experiments on classifying weakly-labeled real-life histopathology images of the breast cancer dataset (\textsc{Breast cancer}) \cite{GBOM:08} and the colon cancer dataset (\textsc{Colon cancer}) \cite{SRTSCR:16}.

\textsc{Breast cancer} consists of 58 weakly labeled $896\times 768$ H\&E images. An image is labeled malignant if it contains breast cancer cells, otherwise it is benign. We divide every image into $32\times 32$ patches. This results in 672 patches per bag. A patch is discarded if it contains 75$\%$ or more of white pixels.

\textsc{Colon cancer} comprises 100 H\&E images. The images originate from a variety of tissue appearance from both normal and malignant regions. For every image the majority of nuclei of each cell were marked. In total there are 22,444 nuclei with associated class label, \textit{i.e.} epithelial, inflammatory, fibroblast, and miscellaneous. A bag is composed of $27\times 27$ patches. Furthermore, a bag is given a positive label if it contains one or more nuclei from the epithelial class. Tagging epithelial cells is highly relevant from a clinical point of view, since colon cancer originates from epithelial cells \cite{RetAl:07}. 

For both datasets we use the model proposed in \cite{SRTSCR:16} for the transformation $f$. All models are trained with the Adam optimization algorithm \cite{KB:14}. Due to the limited amount of data samples in both datasets we performed data augmentation to prevent overfitting. See the Appendix for further details.
\begin{table*}[!htbp]
\small
  \centering
  \caption{Results on \textsc{Breast cancer}. Experiments were run 5 times and an average ($\pm$ a standard error of the mean) is reported.}
 \vspace{2mm}
 \resizebox{0.85\textwidth}{!}{
 \begin{tabular}{lccccc}
    \toprule
    \cmidrule{1-2}
   \textsc{Method}						& \textsc{Accuracy} & \textsc{Precision} & \textsc{Recall} & \textsc{F-score}  & \textsc{AUC}\\
	\midrule   
 Instance+max	& 0.614$\pm$0.020			& 0.585$\pm$0.03			& 0.477$\pm$0.087  			& 		0.506$\pm$0.054 & 0.612$\pm$0.026\\ 
   Instance+mean & 0.672$\pm$0.026			& 0.672$\pm$0.034			& 0.515$\pm$0.056  			& 		0.577$\pm$0.049 & 0.719$\pm$0.019\\
	\midrule   
 Embedding+max	& 0.607$\pm$0.015			& 0.558$\pm$0.013			& 0.546$\pm$0.070  			& 0.543$\pm$0.042 & 0.650$\pm$0.013\\
Embedding+mean	& \textbf{0.741}$\pm$0.023	& \textbf{0.741}$\pm$0.023	& 0.654$\pm$0.054  			& 0.689$\pm$0.034 & \textbf{0.796}$\pm$0.012\\ 
   \midrule  
   Attention		& \textbf{0.745}$\pm$0.018	& 0.718$\pm$0.021			& \textbf{0.715}$\pm$0.046  & \textbf{0.712}$\pm$0.025 & 0.775$\pm$0.016\\ 
   Gated-Attention	& \textbf{0.755}$\pm$0.016	& \textbf{0.728}$\pm$0.016	& \textbf{0.731}$\pm$0.042  & \textbf{0.725}$\pm$0.023 & \textbf{0.799}$\pm$0.020\\
   \bottomrule
 \end{tabular}
 }
 \label{tab:breast_cancer_results}
\end{table*}
\begin{table*}[!htbp]
\small
  \centering
  \caption{Results on \textsc{Colon cancer}. Experiments were run 5 times and an average ($\pm$ a standard error of the mean) is reported.}
  \vspace{2mm}
  \resizebox{0.85\textwidth}{!}{
  \begin{tabular}{lccccc}
    \toprule
    \textsc{Method}						& \textsc{Accuracy} & \textsc{Precision} & \textsc{Recall} & \textsc{F-score}  & \textsc{AUC}\\
	\midrule   
    Instance+max & 0.842 $\pm$ 0.021				& 0.866 $\pm$ 0.017			    & 0.816 $\pm$ 0.031  		    & 0.839 $\pm$ 0.023            & 0.914 $\pm$ 0.010\\ 
    Instance+mean & 0.772 $\pm$ 0.012			    & 0.821 $\pm$ 0.011			    & 0.710 $\pm$ 0.031  		    & 0.759 $\pm$ 0.017            & 0.866 $\pm$ 0.008\\
	\midrule   
    Embedding+max & 0.824 $\pm$ 0.015				& 0.884 $\pm$ 0.014				& 0.753 $\pm$ 0.020  			& 0.813 $\pm$ 0.017				& 0.918 $\pm$ 0.010\\
   Embedding+mean & 0.860 $\pm$ 0.014				& 0.911 $\pm$ 0.011				& 0.804 $\pm$ 0.027  			& 0.853 $\pm$ 0.016 			& 0.940 $\pm$ 0.010\\ 
    \midrule  
    Attention      & \textbf{0.904} $\pm$ 0.011	& \textbf{0.953} $\pm$ 0.014    & \textbf{0.855} $\pm$ 0.017	& \textbf{0.901} $\pm$ 0.011    & \textbf{0.968} $\pm$ 0.009\\ 
    Gated-Attention& \textbf{0.898} $\pm$ 0.020	& \textbf{0.944} $\pm$ 0.016	& \textbf{0.851} $\pm$ 0.035    & \textbf{0.893} $\pm$ 0.022    & \textbf{0.968} $\pm$ 0.010 \\ 
    \bottomrule
  \end{tabular}
  }
  \label{tab:colon_cancer_results}
\end{table*}

\textbf{Results and discussion} We present results in Table \ref{tab:breast_cancer_results} and \ref{tab:colon_cancer_results} for \textsc{Breast cancer} and \textsc{Colon Cancer}, respectively. First, we notice that the obtained results confirm our findings in \textsc{MNIST-bags} experiment that our approach outperforms all other methods. A trend that is especially visible in the small-sample size regime of the \textsc{Mnist-bags}. Surprisingly, the embedding-based method with the $\mathrm{max}$ pooling failed almost completely on \textsc{Breast cancer} but in general this dataset is difficult due to high variability of slides and small number of cases. The proposed method is not only most accurate but it also received the highest recall. High recall is especially important in the medical domain since false negatives could lead to severe consequences including patient fatality. We also notice that the gated-attention mechanism performs better than the plain attention mechanism on \textsc{Breast cancer} while these two behave similarly on \textsc{Colon cancer}.

\begin{figure}[!h]
\vspace{0.2cm}
\centering
\includegraphics[width=.97\columnwidth]{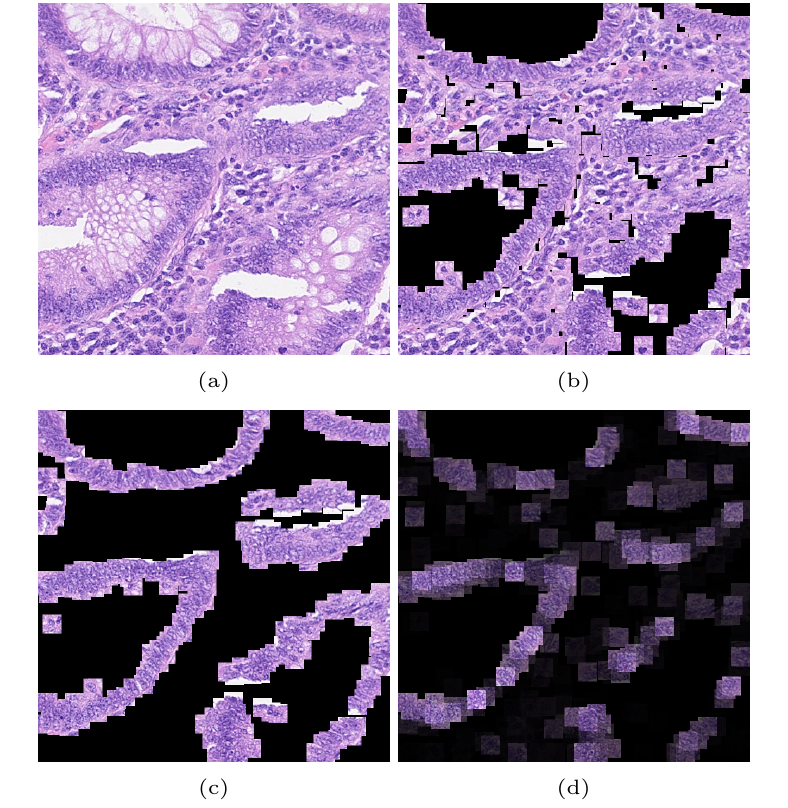}
\caption{(a) H$\&$E stained histology image. (b) 27$\times$27 patches centered around all marked nuclei. (c) Ground truth: Patches that belong to the class epithelial. (d) Heatmap: Every patch from (b) multiplied by its corresponding attention weight, we rescaled the attention weights using $a_k^\prime = (a_k - \min(\textbf{a}))/(\max(\textbf{a}) - \min(\textbf{a}))$.}
\label{fig:colon_attention}
\end{figure}
Eventually, we present the usefulness of the attention mechanism in providing ROIs. In Figure \ref{fig:colon_attention} we show a histopathology image divided into patches containing (mostly) single cells. We create a heatmap by multiplying patches by its corresponding attention weight. Although only image-level annotations are used during training, there is a substantial matching between the heatmap in Figure \ref{fig:colon_attention}(d) and the ground truth in Figure \ref{fig:colon_attention}(c). Additionally, we notice that the instance-based classifier tends to select only a small subset of positive patches (see Figure \ref{fig:colon_cancer_heatmap}(e) in Appendix) that confirms low instance accuracy of the instance-based approach discussed in \cite{KH:15}. For more examples please see the Appendix. 

The obtained results again confirm that the proposed approach attains high predictive performance and allows to properly highlight ROIs. Moreover, the attention weights can be used to create a reliable heatmap.

\section{Conclusion}
\label{conclusion}

In this paper, we proposed a flexible and interpretable MIL approach that is fully parameterized by neural networks. We outlined the usefulness of deep learning for modeling a permutation-invariant bag score function in terms of the Fundamental Theorem of Symmetric Functions. Moreover, we presented a trainable MIL pooling based on the (gated) attention mechanism. We showed empirically on five MIL datasets, one image corpora and two real-life histopathology datasets that our method is on a par with the best performing methods or performs the best in terms of different evaluation metrics. Additionally, we showed that our approach provides an interpretation of the decision by presenting ROIs, which is extremely important in many practical applications.

We strongly believe that the presented line of research is worth pursuing further. Here we focused on a binary MIL problem, however, the multi-class MIL is more interesting and challenging \cite{FZ:17}. Moreover, in some applications it is worth to consider \textit{repulsion points} \cite{SZB:05}, \textit{i.e.}, instances for which a bag is always negative, or assume dependencies among instances within a bag \cite{ZSL:09}. We leave investigating these issues for future research.

\section*{Acknowledgements}

The authors are very grateful to Rianne van den Berg for insightful remarks and discussions.

Maximilian Ilse was funded by the Nederlandse Organisatie voor Wetenschappelijk Onderzoek (Grant ”DLMedIa: Deep Learning for Medical Image Analysis”).

Jakub Tomczak was funded by the European Commission within the Marie Skłodowska-Curie Individual Fellowship (Grant No. 702666, ''Deep learning and Bayesian inference for medical imaging'').


\begin{thebibliography}{46}
\providecommand{\natexlab}[1]{#1}
\providecommand{\url}[1]{\texttt{#1}}
\expandafter\ifx\csname urlstyle\endcsname\relax
  \providecommand{\doi}[1]{doi: #1}\else
  \providecommand{\doi}{doi: \begingroup \urlstyle{rm}\Url}\fi

\bibitem[Andrews et~al.(2003)Andrews, Tsochantaridis, and Hofmann]{ATH:03}
Andrews, Stuart, Tsochantaridis, Ioannis, and Hofmann, Thomas.
\newblock Support vector machines for multiple-instance learning.
\newblock In \emph{NIPS}, pp.\  577--584, 2003.

\bibitem[Bahdanau et~al.(2014)Bahdanau, Cho, and Bengio]{BCB:14}
Bahdanau, Dzmitry, Cho, Kyunghyun, and Bengio, Yoshua.
\newblock Neural machine translation by jointly learning to align and
  translate.
\newblock \emph{arXiv preprint arXiv:1409.0473}, 2014.

\bibitem[Chen et~al.(2006)Chen, Bi, and Wang]{CBW:06}
Chen, Yixin, Bi, Jinbo, and Wang, James~Ze.
\newblock {MILES: Multiple-instance learning via embedded instance selection}.
\newblock \emph{IEEE Transactions on Pattern Analysis and Machine
  Intelligence}, 28\penalty0 (12):\penalty0 1931--1947, 2006.

\bibitem[Cheplygina et~al.(2015{\natexlab{a}})Cheplygina, S{\o}rensen, Tax,
  de~Bruijne, and Loog]{CSTB:17}
Cheplygina, Veronika, S{\o}rensen, Lauge, Tax, David~MJ, de~Bruijne, Marleen,
  and Loog, Marco.
\newblock Label stability in multiple instance learning.
\newblock In \emph{MICCAI}, pp.\  539--546, 2015{\natexlab{a}}.

\bibitem[Cheplygina et~al.(2015{\natexlab{b}})Cheplygina, Tax, and
  Loog]{CTL:15}
Cheplygina, Veronika, Tax, David~MJ, and Loog, Marco.
\newblock Multiple instance learning with bag dissimilarities.
\newblock \emph{Pattern Recognition}, 48\penalty0 (1):\penalty0 264--275,
  2015{\natexlab{b}}.

\bibitem[Dauphin et~al.(2016)Dauphin, Fan, Auli, and Grangier]{DFAG:16}
Dauphin, Yann~N, Fan, Angela, Auli, Michael, and Grangier, David.
\newblock Language modeling with gated convolutional networks.
\newblock \emph{arXiv preprint arXiv:1612.08083}, 2016.

\bibitem[Dietterich et~al.(1997)Dietterich, Lathrop, and
  Lozano-P{\'e}rez]{DLL:97}
Dietterich, Thomas~G, Lathrop, Richard~H, and Lozano-P{\'e}rez, Tom{\'a}s.
\newblock Solving the multiple instance problem with axis-parallel rectangles.
\newblock \emph{Artificial intelligence}, 89\penalty0 (1-2):\penalty0 31--71,
  1997.

\bibitem[Doran \& Ray(2014)Doran and Ray]{DR:14}
Doran, Gary and Ray, Soumya.
\newblock A theoretical and empirical analysis of support vector machine
  methods for multiple-instance classification.
\newblock \emph{Machine Learning}, 97\penalty0 (1-2):\penalty0 79--102, 2014.

\bibitem[Feng \& Zhou(2017)Feng and Zhou]{FZ:17}
Feng, Ji and Zhou, Zhi-Hua.
\newblock {Deep MIML Network}.
\newblock In \emph{AAAI}, pp.\  1884--1890, 2017.

\bibitem[G{\"a}rtner et~al.(2002)G{\"a}rtner, Flach, Kowalczyk, and
  Smola]{GFKS:02}
G{\"a}rtner, Thomas, Flach, Peter~A, Kowalczyk, Adam, and Smola, Alexander~J.
\newblock Multi-instance kernels.
\newblock In \emph{ICML}, volume~2, pp.\  179--186, 2002.

\bibitem[Gelasca et~al.(2008)Gelasca, Byun, Obara, and Manjunath]{GBOM:08}
Gelasca, Elisa~Drelie, Byun, Jiyun, Obara, Boguslaw, and Manjunath, BS.
\newblock Evaluation and benchmark for biological image segmentation.
\newblock In \emph{IEEE International Conference on Image Processing}, pp.\
  1816--1819, 2008.

\bibitem[Glorot \& Bengio(2010)Glorot and Bengio]{GB:10}
Glorot, Xavier and Bengio, Yoshua.
\newblock Understanding the difficulty of training deep feedforward neural
  networks.
\newblock In \emph{AISTATS}, pp.\  249--256, 2010.

\bibitem[Hou et~al.(2016)Hou, Samaras, Kurc, Gao, Davis, and Saltz]{HSKGDS:16}
Hou, Le, Samaras, Dimitris, Kurc, Tahsin~M, Gao, Yi, Davis, James~E, and Saltz,
  Joel~H.
\newblock Patch-based convolutional neural network for whole slide tissue image
  classification.
\newblock In \emph{CVPR}, pp.\  2424--2433, 2016.

\bibitem[Kandemir \& Hamprecht(2015)Kandemir and Hamprecht]{KH:15}
Kandemir, Melih and Hamprecht, Fred~A.
\newblock Computer-aided diagnosis from weak supervision: a benchmarking study.
\newblock \emph{Computerized Medical Imaging and Graphics}, 42:\penalty0
  44--50, 2015.

\bibitem[Kandemir et~al.(2014)Kandemir, Zhang, and Hamprecht]{KZH:14}
Kandemir, Melih, Zhang, Chong, and Hamprecht, Fred~A.
\newblock Empowering multiple instance histopathology cancer diagnosis by cell
  graphs.
\newblock In \emph{MICCAI}, pp.\  228--235, 2014.

\bibitem[Kandemir et~al.(2016)Kandemir, Hau{\ss}mann, Diego, Rajamani, van~der
  Laak, and Hamprecht]{KHDRLH:16}
Kandemir, Melih, Hau{\ss}mann, Manuel, Diego, Ferran, Rajamani, Kumar~T,
  van~der Laak, Jeroen, and Hamprecht, Fred~A.
\newblock {Variational Weakly Supervised Gaussian Processes}.
\newblock In \emph{BMVC}, 2016.

\bibitem[Keeler et~al.(1991)Keeler, Rumelhart, and Leow]{KRL:91}
Keeler, James~D, Rumelhart, David~E, and Leow, Wee~Kheng.
\newblock Integrated segmentation and recognition of hand-printed numerals.
\newblock In \emph{NIPS}, pp.\  557--563, 1991.

\bibitem[Kingma \& Ba(2014)Kingma and Ba]{KB:14}
Kingma, Diederik~P and Ba, Jimmy.
\newblock Adam: A method for stochastic optimization.
\newblock \emph{arXiv preprint arXiv:1412.6980}, 2014.

\bibitem[Kraus et~al.(2016)Kraus, Ba, and Frey]{KBF:16}
Kraus, Oren~Z, Ba, Jimmy~Lei, and Frey, Brendan~J.
\newblock Classifying and segmenting microscopy images with deep multiple
  instance learning.
\newblock \emph{Bioinformatics}, 32\penalty0 (12):\penalty0 i52--i59, 2016.

\bibitem[LeCun et~al.(1998)LeCun, Bottou, Bengio, and Haffner]{LBBH:98}
LeCun, Yann, Bottou, L{\'e}on, Bengio, Yoshua, and Haffner, Patrick.
\newblock Gradient-based learning applied to document recognition.
\newblock \emph{Proceedings of the IEEE}, 86\penalty0 (11):\penalty0
  2278--2324, 1998.

\bibitem[Lin et~al.(2017)Lin, Feng, Santos, Yu, Xiang, Zhou, and
  Bengio]{LFSYXZB:17}
Lin, Zhouhan, Feng, Minwei, Santos, Cicero Nogueira~dos, Yu, Mo, Xiang, Bing,
  Zhou, Bowen, and Bengio, Yoshua.
\newblock A structured self-attentive sentence embedding.
\newblock 2017.

\bibitem[Litjens et~al.(2017)Litjens, Kooi, Bejnordi, Setio, Ciompi,
  Ghafoorian, van~der Laak, van Ginneken, and Sánchez]{LetAl:17}
Litjens, Geert, Kooi, Thijs, Bejnordi, Babak~Ehteshami, Setio, Arnaud
  Arindra~Adiyoso, Ciompi, Francesco, Ghafoorian, Mohsen, van~der Laak,
  Jeroen~A.W.M., van Ginneken, Bram, and Sánchez, Clara~I.
\newblock A survey on deep learning in medical image analysis.
\newblock \emph{Medical Image Analysis}, 42:\penalty0 60 -- 88, 2017.

\bibitem[Liu et~al.(2012)Liu, Wu, and Zhou]{LIU:12}
Liu, Guoqing, Wu, Jianxin, and Zhou, Zhi-Hua.
\newblock Key instance detection in multi-instance learning.
\newblock In \emph{JMLR}, volume~25, pp.\  253--268, 2012.

\bibitem[Maron \& Lozano-P{\'e}rez(1998)Maron and Lozano-P{\'e}rez]{ML:98}
Maron, Oded and Lozano-P{\'e}rez, Tom{\'a}s.
\newblock A framework for multiple-instance learning.
\newblock In \emph{NIPS}, pp.\  570--576, 1998.

\bibitem[Oquab et~al.(2014)Oquab, Bottou, Laptev, Sivic, et~al.]{OBLS:14}
Oquab, Maxime, Bottou, L{\'e}on, Laptev, Ivan, Sivic, Josef, et~al.
\newblock Weakly supervised object recognition with convolutional neural
  networks.
\newblock In \emph{NIPS}, 2014.

\bibitem[Pappas \& Popescu-Belis(2014)Pappas and Popescu-Belis]{PP:14}
Pappas, Nikolaos and Popescu-Belis, Andrei.
\newblock {Explaining the stars: Weighted multiple-instance learning for
  aspect-based sentiment analysis}.
\newblock In \emph{EMNLP}, pp.\  455--466, 2014.

\bibitem[Pappas \& Popescu-Belis(2017)Pappas and Popescu-Belis]{PP:17}
Pappas, Nikolaos and Popescu-Belis, Andrei.
\newblock {Explicit Document Modeling through Weighted Multiple-Instance
  Learning}.
\newblock \emph{Journal of Artificial Intelligence Research}, 58:\penalty0
  591--626, 2017.

\bibitem[Pinheiro \& Collobert(2015)Pinheiro and Collobert]{PC:15}
Pinheiro, Pedro~O and Collobert, Ronan.
\newblock From image-level to pixel-level labeling with convolutional networks.
\newblock In \emph{CVPR}, pp.\  1713--1721, 2015.

\bibitem[Qi et~al.(2017)Qi, Su, Mo, and Guibas]{QSMG:17}
Qi, Charles~R, Su, Hao, Mo, Kaichun, and Guibas, Leonidas~J.
\newblock {PointNet: Deep learning on point sets for 3d classification and
  segmentation}.
\newblock In \emph{CVPR}, 2017.

\bibitem[Quellec et~al.(2017)Quellec, Cazuguel, Cochener, and Lamard]{QCCL:17}
Quellec, Gwenole, Cazuguel, Guy, Cochener, Beatrice, and Lamard, Mathieu.
\newblock Multiple-instance learning for medical image and video analysis.
\newblock \emph{IEEE Reviews in Biomedical Engineering}, 2017.

\bibitem[Raffel \& Ellis(2015)Raffel and Ellis]{RE:15}
Raffel, Colin and Ellis, Daniel~PW.
\newblock Feed-forward networks with attention can solve some long-term memory
  problems.
\newblock 2015.

\bibitem[Ramon \& De~Raedt(2000)Ramon and De~Raedt]{RD:00}
Ramon, Jan and De~Raedt, Luc.
\newblock Multi instance neural networks.
\newblock In \emph{ICML Workshop on Attribute-value and Relational Learning},
  pp.\  53--60, 2000.

\bibitem[Raykar et~al.(2008)Raykar, Krishnapuram, Bi, Dundar, and
  Rao]{RKBDR:08}
Raykar, Vikas~C, Krishnapuram, Balaji, Bi, Jinbo, Dundar, Murat, and Rao,
  R~Bharat.
\newblock Bayesian multiple instance learning: automatic feature selection and
  inductive transfer.
\newblock In \emph{ICML}, pp.\  808--815, 2008.

\bibitem[Ricci-Vitiani et~al.(2007)Ricci-Vitiani, Lombardi, Pilozzi, Biffoni,
  Todaro, Peschle, and De~Maria]{RetAl:07}
Ricci-Vitiani, Lucia, Lombardi, Dario~G, Pilozzi, Emanuela, Biffoni, Mauro,
  Todaro, Matilde, Peschle, Cesare, and De~Maria, Ruggero.
\newblock Identification and expansion of human colon-cancer-initiating cells.
\newblock \emph{Nature}, 445\penalty0 (7123):\penalty0 111, 2007.

\bibitem[Ruifrok \& Johnston(2001)Ruifrok and Johnston]{RJ:01}
Ruifrok, Arnout~C and Johnston, Dennis~A.
\newblock Quantification of histochemical staining by color deconvolution.
\newblock \emph{Analytical and Quantitative Cytology and Histology},
  23\penalty0 (4):\penalty0 291--299, 2001.

\bibitem[Scott et~al.(2005)Scott, Zhang, and Brown]{SZB:05}
Scott, Stephen, Zhang, Jun, and Brown, Joshua.
\newblock On generalized multiple-instance learning.
\newblock \emph{International Journal of Computational Intelligence and
  Applications}, 5\penalty0 (01):\penalty0 21--35, 2005.

\bibitem[Sirinukunwattana et~al.(2016)Sirinukunwattana, Raza, Tsang, Snead,
  Cree, and Rajpoot]{SRTSCR:16}
Sirinukunwattana, Korsuk, Raza, Shan E~Ahmed, Tsang, Yee-Wah, Snead, David~RJ,
  Cree, Ian~A, and Rajpoot, Nasir~M.
\newblock Locality sensitive deep learning for detection and classification of
  nuclei in routine colon cancer histology images.
\newblock \emph{IEEE Transactions on Medical Imaging}, 35\penalty0
  (5):\penalty0 1196--1206, 2016.

\bibitem[Wang et~al.(2017)Wang, Jiang, Qian, Yang, Li, Zhang, Wang, and
  Tang]{WJQYLZWT:17}
Wang, Fei, Jiang, Mengqing, Qian, Chen, Yang, Shuo, Li, Cheng, Zhang, Honggang,
  Wang, Xiaogang, and Tang, Xiaoou.
\newblock {Residual Attention Network for Image Classification}.
\newblock In \emph{CVPR}, 2017.

\bibitem[Wang et~al.(2016)Wang, Yan, Tang, Bai, and Liu]{WYTPBL:18}
Wang, Xinggang, Yan, Yongluan, Tang, Peng, Bai, Xiang, and Liu, Wenyu.
\newblock Revisiting multiple instance neural networks.
\newblock \emph{Pattern Recognition}, 74:\penalty0 15--24, 2016.

\bibitem[Wei et~al.(2017)Wei, Wu, and Zhou]{WWZ:17}
Wei, Xiu-Shen, Wu, Jianxin, and Zhou, Zhi-Hua.
\newblock Scalable algorithms for multi-instance learning.
\newblock \emph{IEEE Transactions on Neural Networks and Learning Systems},
  28\penalty0 (4):\penalty0 975--987, 2017.

\bibitem[Xu et~al.(2015)Xu, Ba, Kiros, Cho, Courville, Salakhudinov, Zemel, and
  Bengio]{XBKCCSZB:15}
Xu, Kelvin, Ba, Jimmy, Kiros, Ryan, Cho, Kyunghyun, Courville, Aaron,
  Salakhudinov, Ruslan, Zemel, Rich, and Bengio, Yoshua.
\newblock Show, attend and tell: Neural image caption generation with visual
  attention.
\newblock In \emph{ICML}, pp.\  2048--2057, 2015.

\bibitem[Zaheer et~al.(2017)Zaheer, Kottur, Ravanbakhsh, Poczos, Salakhutdinov,
  and Smola]{ZKRPSS:17}
Zaheer, Manzil, Kottur, Satwik, Ravanbakhsh, Siamak, Poczos, Barnabas,
  Salakhutdinov, Ruslan, and Smola, Alexander.
\newblock {Deep Sets}.
\newblock In \emph{NIPS}. 2017.

\bibitem[Zhang et~al.(2006)Zhang, Platt, and Viola]{ZPV:06}
Zhang, Cha, Platt, John~C, and Viola, Paul~A.
\newblock Multiple instance boosting for object detection.
\newblock In \emph{NIPS}, pp.\  1417--1424, 2006.

\bibitem[Zhang \& Goldman(2002)Zhang and Goldman]{ZG:02}
Zhang, Qi and Goldman, Sally~A.
\newblock Em-dd: An improved multiple-instance learning technique.
\newblock In \emph{NIPS}, pp.\  1073--1080, 2002.

\bibitem[Zhou et~al.(2009)Zhou, Sun, and Li]{ZSL:09}
Zhou, Zhi-Hua, Sun, Yu-Yin, and Li, Yu-Feng.
\newblock Multi-instance learning by treating instances as non-iid samples.
\newblock In \emph{ICML}, pp.\  1249--1256, 2009.

\bibitem[Zhu et~al.(2017)Zhu, Lou, Vang, and Xie]{ZLVX:17}
Zhu, Wentao, Lou, Qi, Vang, Yeeleng~Scott, and Xie, Xiaohui.
\newblock Deep multi-instance networks with sparse label assignment for whole
  mammogram classification.
\newblock In \emph{MICCAI}, pp.\  603--611, 2017.

\end{thebibliography}


\newpage
\onecolumn

\section{Appendix}
\label{appendix}

\subsection{Deep MIL approaches}
\label{app:deep_MIL}

\begin{figure}[!thbp]
	\centering
	\includegraphics[width=\columnwidth]{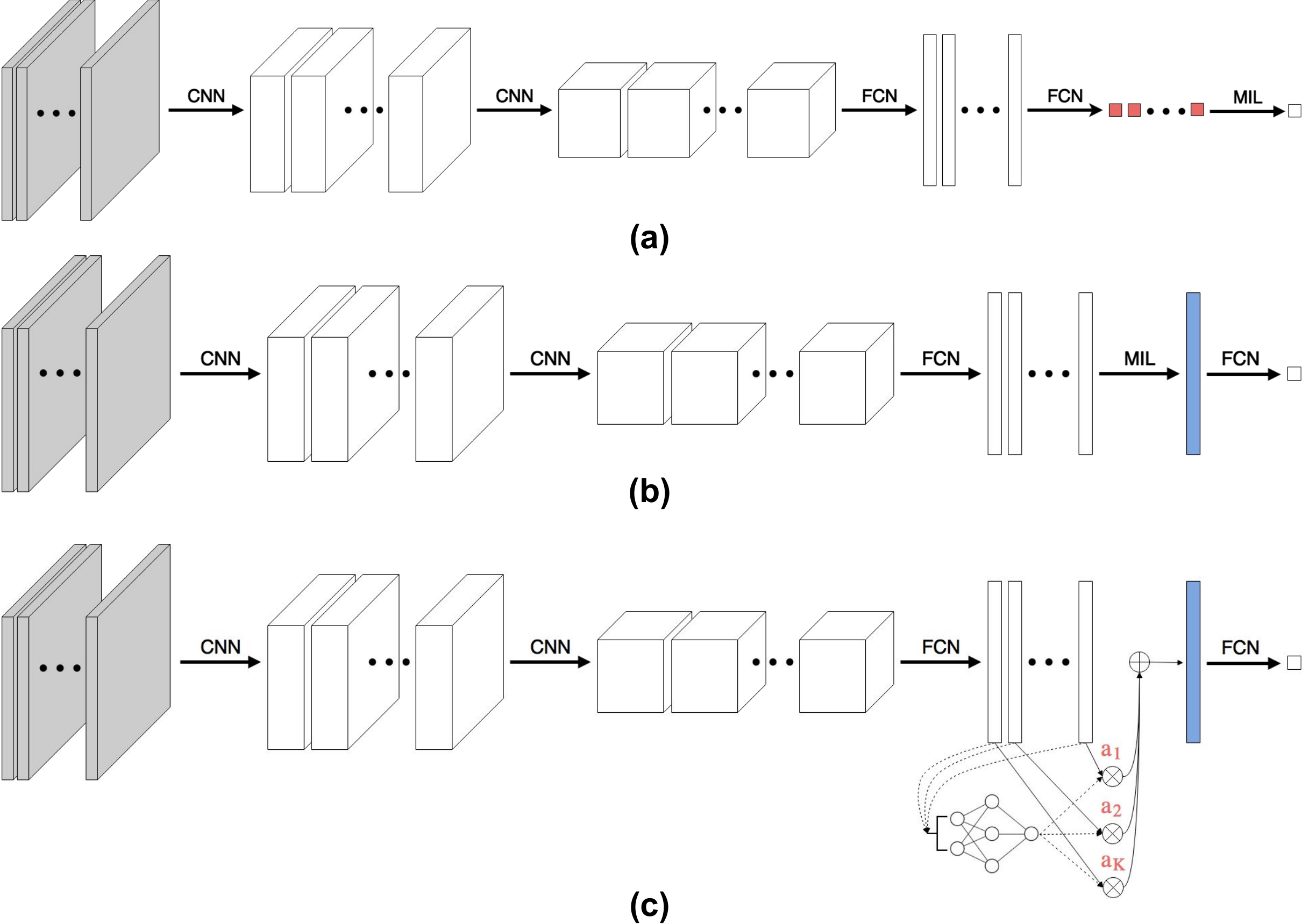}
    \vskip -2mm
	\caption{Deep MIL approaches: (a) the instance-based approach, (b) the embedding-based approach, (c) the proposed approach with the attention mechanism as the MIL pooling. Red color corresponds to instance scores, blue color depicts a bag vector representation. \textit{Best viewed in color.}}
	\label{fig:architectures}
\end{figure}

In Figure \ref{fig:architectures} we present three deep MIL approaches discussed in the paper.

\subsection{Code}

The implementation of our methods is available online at \url{https://github.com/AMLab-Amsterdam/AttentionDeepMIL}. All experiments were run on NVIDIA TITAN X Pascal with a batch size of 1 (= 1 bag) for all datasets.

\subsection{Classical MIL datasets}
\label{app:classic}

\paragraph{Additional details} In Table \ref{tab:classic_MIL_results} a general description of the five benchmark MIL datasets used in the experiments is given. In Tables \ref{tab:classic_mil_embedding_model} and \ref{tab:classic_mil_instance_model} we present architectures of the embedding-based and the instance-based models, respectively. We denote a fully-connected layer by 'fc' and the number of output hidden units is provided after a dash. The ReLU non-linearity was used. In Table \ref{tab:classic_mil_optimizer} the details of the optimization (learning) procedure are given. We provide values of hyperparameters determined by the model selection procedure for which the highest validation performance was achieved.

\begin{table}[!htbp]
  \centering
  \caption{Overview of classical MIL datasets.}
  \vspace{2mm}
  \begin{tabular}{|c|c|c|c|}
    \hline
    Dataset & $\#$ of bags & $\#$ of instances & $\#$ of features\\
    \hline 
    Musk1 & 92 & 476 & 166 \\
    Musk2 & 102 & 6598 & 166 \\
    Tiger & 200 & 1220 & 230 \\
    Fox & 200 & 1302 & 230 \\
    Elephant & 200 & 1391 & 230\\
    \hline
  \end{tabular}
  \label{tab:classic_mil_overview}
\end{table}

\begin{minipage}{.5\textwidth}
\begin{table}[H]
  \centering
  \caption{Classical MIL datasets: The embedding-based model architecture \cite{WYTPBL:18}.}
  \vspace{2mm}
  \begin{tabular}{|c|c|}
    \hline
    Layer & Type\\
    \hline   
    1 & fc-256 + $\mathrm{ReLU}$\\
    2 & dropout \\
    3 & fc-128 + $\mathrm{ReLU}$\\
    4 & dropout \\
    5 & fc-64 + $\mathrm{ReLU}$\\
    6 & dropout \\
    7 & mil-$\mathrm{max}$/mil-$\mathrm{mean}$/mil-attention-64\\
    8 & fc-1 + $\mathrm{sigm}$\\
    \hline
  \end{tabular}
  \label{tab:classic_mil_embedding_model}
\end{table}
\end{minipage}
\hspace{5pt}
\begin{minipage}{.5\textwidth}
\begin{table}[H]
  \centering
  \caption{Classical MIL datasets: The instance-based model architecture \cite{WYTPBL:18}.}
  \vspace{2mm}
  \begin{tabular}{|c|c|}
    \hline
    Layer & Type\\
    \hline   
    1 & fc-256 + $\mathrm{ReLU}$\\
    2 & dropout \\
    3 & fc-128 + $\mathrm{ReLU}$\\
    4 & dropout \\
    5 & fc-64 + $\mathrm{ReLU}$\\
    6 & dropout \\
    7 & fc-1 + $\mathrm{sigm}$\\
    8 & mil-$\mathrm{max}$/mil-$\mathrm{mean}$\\
    \hline
  \end{tabular}
  \label{tab:classic_mil_instance_model}
\end{table}
\end{minipage}

\begin{table*}[!hbpt]
  \centering
  \caption{Classical MIL datasets: The optimization procedure details \cite{WYTPBL:18}.}
  \vspace{2mm}
  \begin{tabular}{|c|c|c|c|c|c|c|}
    \hline
    Experiment & Optimizer & Momentum & Learning rate & Weight decay & Epochs & Stopping criteria\\
    \hline
    Musk1 & SGD & 0.9 & 0.0005 & 0.005 & 100 & lowest validation error and loss\\
    Musk2 & SGD & 0.9 & 0.0005 & 0.03 & 100 & lowest validation error and loss\\
    Tiger & SGD & 0.9 & 0.0001 & 0.01 & 100 & lowest validation error and loss\\
    Fox & SGD & 0.9 & 0.0005 & 0.005 & 100 & lowest validation error and loss\\
    Elephant & SGD & 0.9 & 0.0001 & 0.005 & 100 & lowest validation error and loss\\
    \hline
  \end{tabular}
  \label{tab:classic_mil_optimizer}
\end{table*}

\subsection{MNIST-bags}
\label{app:mnist}

\paragraph{Additional details} In Tables \ref{tab:mnist_bag_embedding_model} and \ref{tab:mnist_bag_instance_model} we present architectures of the embedding-based and the instance-based models for \textsc{Mnist-bags}, respectively. We denote a convolutional layer by 'conv', in brackets we provide kernel size, stride and padding, and the number of kernels is provided after a dash. The convolutional max-pooling layer is denoted by 'maxpool' and the pooling size is given in brackets. The ReLU non-linearity was used. In Table \ref{tab:mnist_bag_optimizer} the details of the optimization (learning) procedure for deep MIL approach are given. The details of the SVM are given in Table \ref{tab:svm_config}. We provide values of hyperparameters determined by the model selection procedure for which the highest validation performance was achieved.

\begin{minipage}{.5\textwidth}
\begin{table}[H]
  \centering
  \caption{MNIST-bags: The embedding-based model architecture \cite{LBBH:98}.}
  \vspace{2mm}
  \begin{tabular}{|c|c|}
    \hline
    Layer & Type\\
    \hline 
    1 & conv(5,1,0)-20 + $\mathrm{ReLU}$\\
    2 & maxpool(2,2) \\
    3 & conv(5,1,0)-50 + $\mathrm{ReLU}$\\
    4 & maxpool(2,2) \\
    5 & fc-500 + $\mathrm{ReLU}$\\
    6 & mil-$\mathrm{max}$/mil-$\mathrm{mean}$/mil-attention-128\\
    7 & fc-1 + $\mathrm{sigm}$\\
    \hline
  \end{tabular}
  \label{tab:mnist_bag_embedding_model}
\end{table}
\end{minipage}
\hspace{5pt}
\begin{minipage}{.5\textwidth}
\begin{table}[H]
  \centering
  \caption{MNIST-bags: The instance-based model architecture \cite{LBBH:98}.}
  \vspace{2mm}
  \begin{tabular}{|c|c|}
    \hline
    Layer & Type\\
    \hline 
    1 & conv(5,1,0)-20 + $\mathrm{ReLU}$\\
    2 & maxpool(2,2) \\
    3 & conv(5,1,0)-50 + $\mathrm{ReLU}$\\
    4 & maxpool(2,2) \\
    5 & fc-500 + $\mathrm{ReLU}$\\
    6 & fc-1 + $\mathrm{sigm}$\\
    7 & mil-$\mathrm{max}$/mil-$\mathrm{mean}$\\
    \hline
  \end{tabular}
  \label{tab:mnist_bag_instance_model}
\end{table}
\end{minipage}

\begin{table*}[!hbpt]
  \centering
  \caption{MNIST-bags: The optimization procedure details.}
  \vspace{2mm}
  \begin{tabular}{|c|c|c|c|c|c|c|}
    \hline
    Experiment & Optimizer & $\beta_1,$ $\beta_2$ & Learning rate & Weight decay & Epochs & Stopping criteria\\
    \hline
    All & Adam & 0.9, 0.999 & 0.0005 & 0.0001 & 200 & lowest validation error+loss\\
    \hline
  \end{tabular}
  \label{tab:mnist_bag_optimizer}
\end{table*}
\begin{table*}[!hbpt]
  \centering
  \caption{MNIST-bags: SVM configuration.}
  \vspace{2mm}
  \begin{tabular}{|c|c|c|c|c|c|}
    \hline
    Model & Features & Kernel & $C$ & $\gamma$ & Max iterations\\
    \hline
    MI-SVM & Raw pixel values & RBF & 5 & 0.0005 & 200\\
    \hline
  \end{tabular}
  \label{tab:svm_config}
\end{table*}

\paragraph{Additional results} In Tables \ref{tab:mnist_mean_10}, \ref{tab:mnist_mean_50} and \ref{tab:mnist_mean_100} we present the test AUC value for 10, 50 and 100 instances on average per a bag, respectively.

\begin{figure*}[!hbpt]
\centering
\includegraphics[width=\columnwidth]{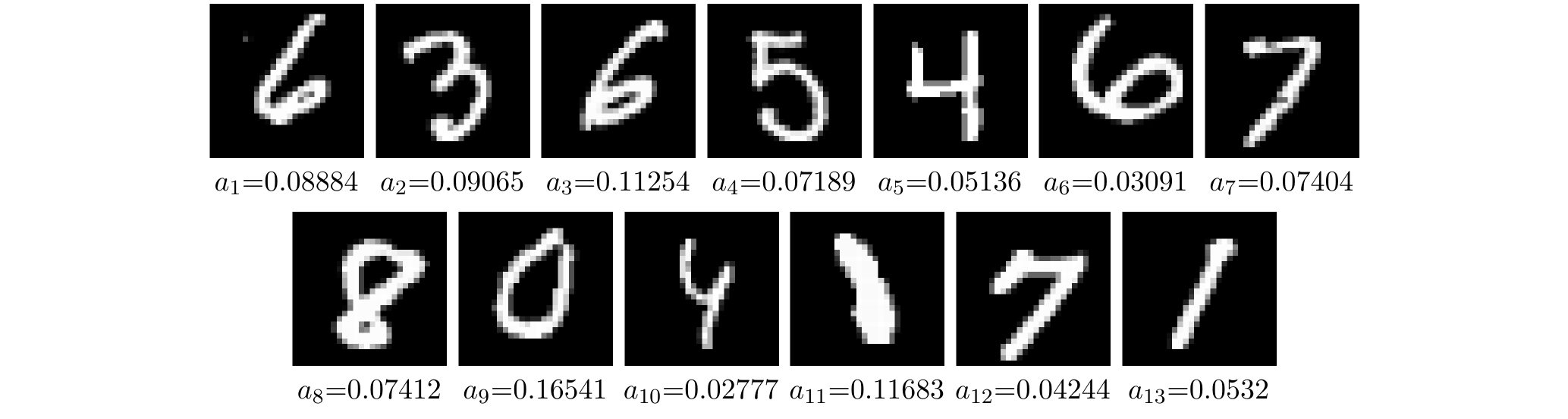}
\vskip -5pt
\caption{Example of attention weights for a negative bag.}
\label{fig:mnist_attention_negative}
\end{figure*}

\begin{figure*}[!hbpt]
\centering
\includegraphics[width=\columnwidth]{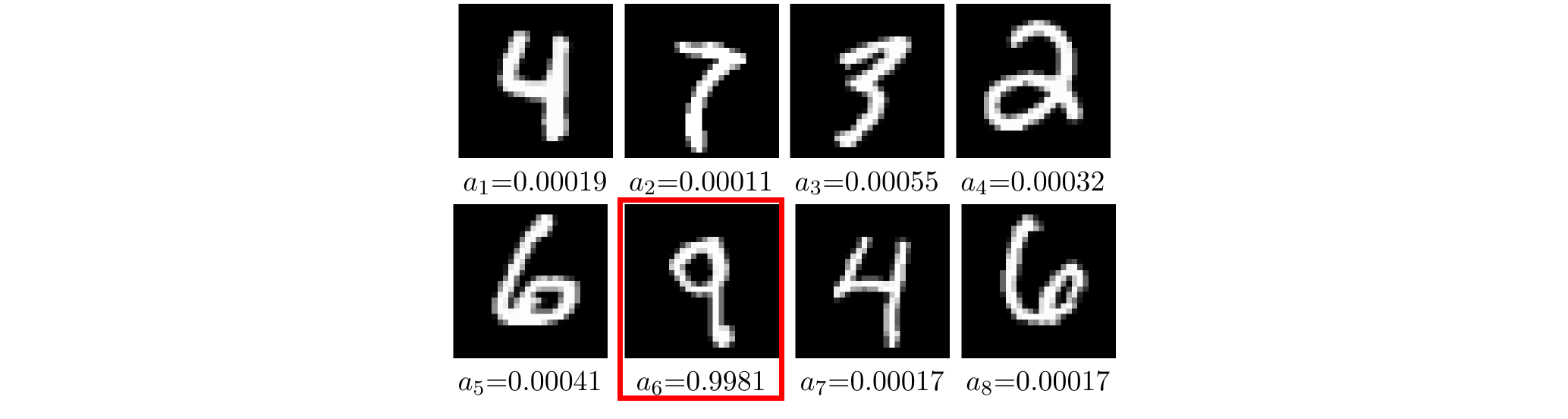}
\vskip -5pt
\caption{Example of attention weights for a positive bag containing a single '9'.}
\label{fig:mnist_attention_positive_1}
\end{figure*}

\begin{figure*}[!hbpt]
\centering
\includegraphics[width=\columnwidth]{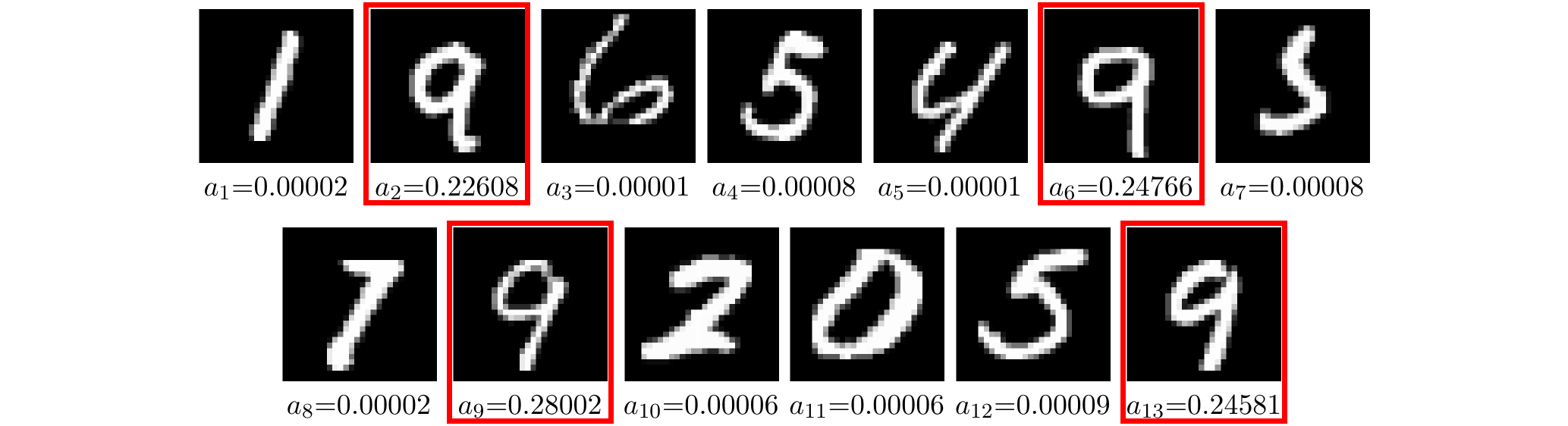}
\vskip -5pt
\caption{Example of attention weights for a positive bag containing multiple '9's.}
\label{fig:mnist_attention_positive_2}
\end{figure*}

In Figure \ref{fig:mnist_attention_negative} a negative bag is presented. In Figure \ref{fig:mnist_attention_positive_1} a positive bag with a single '9' is given. In Figure \ref{fig:mnist_attention_positive_2} a positive bag with multiple '9's is presented. In all figures attention weights are provided and in the case of positive bags a red rectangle highlights positive instances.

\begin{table*}[!hbpt]
\small
  \centering
  \caption{The test AUC for \textsc{MNIST-bags} with on average $10$ instances per bag for different numbers of training bags.}
  \vspace{2mm}
  \begin{tabular}{|c|c|c|c|c|c|c|c|}
    \hline
    $\#$ of training bags & 50 & 100 & 150 & 200 & 300 & 400 & 500\\
	\hline
    Instance+max & 0.553 $\pm$ 0.053 & 0.745 $\pm$ 0.100 & 0.960 $\pm$ 0.004 & 0.979 $\pm$ 0.001&0.984 $\pm$ 0.001 & 0.986 $\pm$ 0.001 & 0.986 $\pm$ 0.001\\ 
    Instance+mean & 0.663 $\pm$ 0.014 & 0.676 $\pm$ 0.012 & 0.694 $\pm$ 0.010 & 0.694 $\pm$ 0.017& 0.709 $\pm$ 0.020& 0.693 $\pm$ 0.023&0.712 $\pm$ 0.018\\
	\hline   
    MI-SVM & 0.697 $\pm$ 0.054 & 0.851 $\pm$ 0.009& 0.862 $\pm$ 0.008& 0.898 $\pm$ 0.014& 0.926 $\pm$ 0.004& 0.942 $\pm$ 0.002& 0.948 $\pm$ 0.002\\
    Embedded+max & 0.713 $\pm$ 0.016 & 0.914 $\pm$ 0.011& 0.954 $\pm$ 0.005& 0.968 $\pm$ 0.001& 0.980 $\pm$ 0.001& 0.981 $\pm$ 0.003& 0.986 $\pm$ 0.002\\
    Embedded+mean & 0.695 $\pm$ 0.026 & 0.841 $\pm$ 0.027& 0.926 $\pm$ 0.004& 0.953 $\pm$ 0.004& 0.974 $\pm$ 0.002& 0.980 $\pm$ 0.001&0.984 $\pm$ 0.002\\ 
    \hline
    Attention & 0.768 $\pm$ 0.054 & 0.948 $\pm$ 0.007& 0.949 $\pm$ 0.006& 0.970 $\pm$ 0.003& 0.980 $\pm$ 0.000& 0.982 $\pm$ 0.001& 0.986 $\pm$ 0.001\\ 
    Gated Attention & 0.753 $\pm$ 0.054 & 0.916 $\pm$ 0.013& 0.955 $\pm$ 0.003& 0.974 $\pm$ 0.002& 0.980 $\pm$ 0.004& 0.983 $\pm$ 0.002& 0.987 $\pm$ 0.001\\
   \hline
  \end{tabular}
  \label{tab:mnist_mean_10}
\end{table*}

\begin{table*}[!hbpt]
\small
  \centering
  \caption{The test AUC for \textsc{MNIST-bags} with on average $50$ instances per bag for different numbers of training bags.}
  \vspace{2mm}
  \begin{tabular}{|c|c|c|c|c|c|c|c|}
  \hline
    $\#$ of training bags & 50 & 100 & 150 & 200 & 300 & 400 & 500\\
	\hline
    Instance+max & 0.576 $\pm$ 0.059& 0.715 $\pm$ 0.096& 0.937 $\pm$ 0.045& 0.992 $\pm$ 0.002& 0.994 $\pm$ 0.001& 0.997 $\pm$ 0.001&0.997 $\pm$ 0.001\\ 
    Instance+mean & 0.737 $\pm$ 0.014& 0.744 $\pm$ 0.029& 0.824 $\pm$ 0.012& 0.813 $\pm$ 0.030& 0.722 $\pm$ 0.021& 0.728 $\pm$ 0.017&0.798 $\pm$ 0.011\\
	\hline   
    MI-SVM & 0.824 $\pm$ 0.067& 0.946 $\pm$ 0.004& 0.959 $\pm$ 0.002& 0.967 $\pm$ 0.002& 0.975 $\pm$ 0.001& 0.976 $\pm$ 0.001&0.979 $\pm$ 0.001\\
    Embedded+max & 0.872 $\pm$ 0.039& 0.984 $\pm$ 0.005& 0.992 $\pm$ 0.001& 0.996 $\pm$ 0.001& 0.996 $\pm$ 0.001& 0.997 $\pm$ 0.001&0.997 $\pm$ 0.001\\
    Embedded+mean & 0.841 $\pm$ 0.013& 0.906 $\pm$ 0.046& 0.983 $\pm$ 0.005& 0.992 $\pm$ 0.001& 0.996 $\pm$ 0.001&0.997 $\pm$ 0.001&0.997 $\pm$ 0.001\\ 
    \hline
    Attention & 0.967 $\pm$ 0.010& 0.982 $\pm$ 0.003& 0.990 $\pm$ 0.002& 0.993 $\pm$ 0.002& 0.989 $\pm$ 0.003& 0.994 $\pm$ 0.001&0.995 $\pm$ 0.001\\ 
    Gated Attention & 0.920 $\pm$ 0.042& 0.977 $\pm$ 0.006& 0.993 $\pm$ 0.003& 0.991 $\pm$ 0.002& 0.994 $\pm$ 0.002& 0.995 $\pm$ 0.001&0.996 $\pm$ 0.001\\ 
    \hline
  \end{tabular}
  \label{tab:mnist_mean_50}
\end{table*}

\begin{table*}[!hbpt]
\small
  \centering
  \caption{The test AUC for \textsc{MNIST-bags} with on average $100$ instances per bag for different numbers of training bags.}
  \vspace{2mm}
  \begin{tabular}{|c|c|c|c|c|c|c|c|}
    \hline
    $\#$ of training bags & 50 & 100 & 150 & 200 & 300 & 400 & 500\\
	\hline
    Instance+max & 0.543 $\pm$ 0.054& 0.804 $\pm$ 0.107& 0.899 $\pm$ 0.086& 0.999 $\pm$ 0.000& 1.000 $\pm$ 0.000& 1.000 $\pm$ 0.000& 1.000 $\pm$ 0.000\\ 
    Instance+mean & 0.842 $\pm$ 0.023& 0.855 $\pm$ 0.025& 0.824 $\pm$ 0.014&0.896 $\pm$ 0.037& 0.859 $\pm$ 0.029& 0.899 $\pm$ 0.012& 0.868 $\pm$ 0.016\\
	\hline   
    MI-SVM & 0.871 $\pm$ 0.060& 0.991 $\pm$ 0.002& 0.994 $\pm$ 0.002& 0.996 $\pm$ 0.001& 0.997 $\pm$ 0.001& 0.998 $\pm$ 0.001&0.998 $\pm$ 0.001\\
    Embedded+max & 0.977 $\pm$ 0.009& 0.999 $\pm$ 0.001& 1.000 $\pm$ 0.000& 1.000 $\pm$ 0.000& 1.000 $\pm$ 0.000& 1.000 $\pm$ 0.000&1.000 $\pm$ 0.000\\
    Embedded+mean & 0.959 $\pm$ 0.010& 0.990 $\pm$ 0.003& 0.998 $\pm$ 0.001& 0.900 $\pm$ 0.089& 1.000 $\pm$ 0.000& 1.000 $\pm$ 0.000&1.000 $\pm$ 0.000\\ 
    \hline
    Attention & 0.996 $\pm$ 0.001& 0.998 $\pm$ 0.001& 0.999 $\pm$ 0.000& 0.998 $\pm$ 0.001& 1.000 $\pm$ 0.000& 1.000 $\pm$ 0.000&1.000 $\pm$ 0.000\\ 
    Gated Attention & 0.998 $\pm$ 0.001& 0.999 $\pm$ 0.000& 0.998 $\pm$ 0.001& 0.998 $\pm$ 0.001& 0.999 $\pm$ 0.000& 1.000 $\pm$ 0.000&1.000 $\pm$ 0.000\\ 
    \hline
  \end{tabular}
  \label{tab:mnist_mean_100}
\end{table*}

\fboxsep=0pt
\fboxrule=2pt

\newpage
\subsection{Histopathology datasets}
\label{app:histopathology}

\paragraph{Data augmentation} We randomly adjust the amount of H$\&$E by decomposing the RGB color of the tissue into the H$\&$E color space \cite{RJ:01}, followed by multiplying the magnitude of H$\&$E for a pixel by two i.i.d. Gaussian random variables with expectation equal to one. We randomly rotate and mirror every patch. Lastly, we perform color normalization on every patch.

\paragraph{Additional details} In Tables \ref{tab:histo_embedded_model} and \ref{tab:histo_instance_model} we present architectures of the embedding-based and the instance-based models for histopathology datasets, respectively. In Table \ref{tab:histo_optimizer} the details of the optimization (learning) procedure for deep MIL approach are given. We provide values of hyperparameters determined by the model selection procedure for which the highest validation performance was achieved.

\begin{minipage}{.5\textwidth}
\begin{table}[H]
  \centering
  \caption{Histopathology: The embedding-based model architecture  \cite{SRTSCR:16}.}
  \vspace{2mm}
  \begin{tabular}{|c|c|}
    \hline
    Layer & Type\\
    \hline  
    1 & conv(4,1,0)-36 + $\mathrm{ReLU}$\\
    2 & maxpool(2,2) \\
    3 & conv(3,1,0)-48 + $\mathrm{ReLU}$\\
    4 & maxpool(2,2) \\
    5 & fc-512 + $\mathrm{ReLU}$\\
    6 & dropout \\
    7 & fc-512 + $\mathrm{ReLU}$\\
    8 & dropout \\
    9 & mil-$\mathrm{max}$/mil-$\mathrm{mean}$/mil-attention-128\\
    10 & fc-1 + $\mathrm{sigm}$\\
    \hline
  \end{tabular}
  \label{tab:histo_embedded_model}
\end{table}
\end{minipage}
\hspace{5pt}
\begin{minipage}{.5\textwidth}
\begin{table}[H]
  \centering
  \caption{Histopathology: The instance-based model architecture  \cite{SRTSCR:16}.}
  \vspace{2mm}
  \begin{tabular}{|c|c|}
    \hline
    Layer & Type\\
    \hline  
    1 & conv(4,1,0)-36 + $\mathrm{ReLU}$\\
    2 & maxpool(2,2) \\
    3 & conv(3,1,0)-48 + $\mathrm{ReLU}$\\
    4 & maxpool(2,2) \\
    5 & fc-512 + $\mathrm{ReLU}$\\
    6 & dropout \\
    7 & fc-512 + $\mathrm{ReLU}$\\
    8 & dropout \\
    9 & fc-1 + $\mathrm{sigm}$\\
    10 & mil-$\mathrm{max}$/mil-$\mathrm{mean}$ \\
    \hline
  \end{tabular}
  \label{tab:histo_instance_model}
\end{table}
\end{minipage}

\begin{table*}[!hbpt]
  \centering
  \caption{Histopathology: The optimization procedure details.}
  \vspace{2mm}
  \begin{tabular}{|c|c|c|c|c|c|c|}
    \hline
    Experiment & Optimizer & $\beta_1,$ $\beta_2$ & Learning rate & Weight decay & Epochs & Stopping criteria\\
    \hline
    All & Adam & 0.9, 0.999 & 0.0001 & 0.0005 & 100 & lowest validation error+loss\\
    \hline
  \end{tabular}
  \label{tab:histo_optimizer}
\end{table*}

\begin{figure*}[!hbpt]
\centering
\includegraphics[width=\columnwidth]{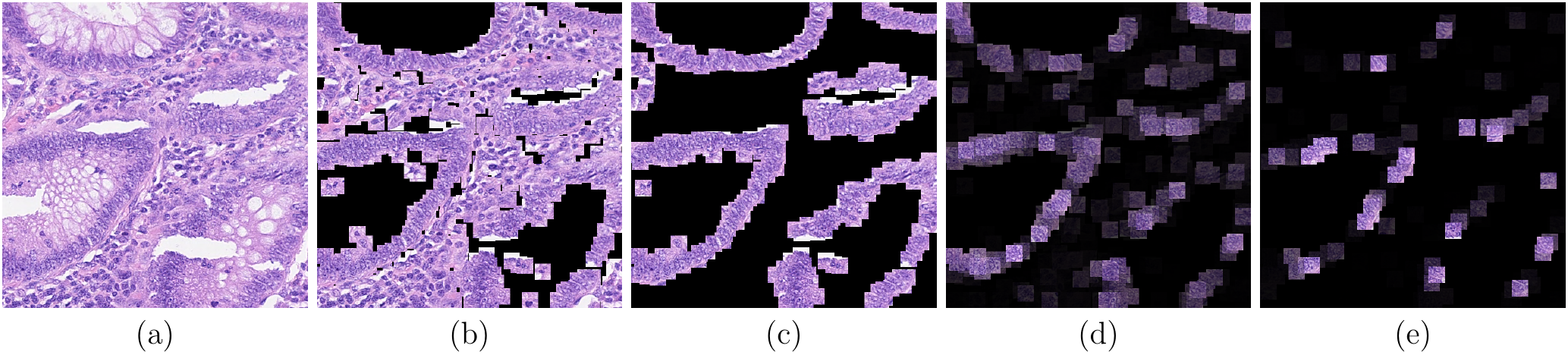}
\vskip -5pt
\caption{Colon cancer example 1: (a) H$\&$E stained histology image. (b) 27$\times$27 patches centered around all marked nuclei. (c) Ground truth: Patches that belong to the class epithelial. (d) Attention heatmap: Every patch from (b) multiplied by its attention weight. (e) Instance+$\mathrm{max}$ heatmap: Every patch from (b) multiplied by its score from the \textsc{instance}+$\max$ model. We rescaled the attention weights and instance scores using $a_k^\prime = (a_k - \min(\textbf{a}))/(\max(\textbf{a}) - \min(\textbf{a}))$.}
\label{fig:colon_cancer_heatmap}
\end{figure*}

\begin{figure*}[!hbpt]
\centering
\includegraphics[width=\columnwidth]{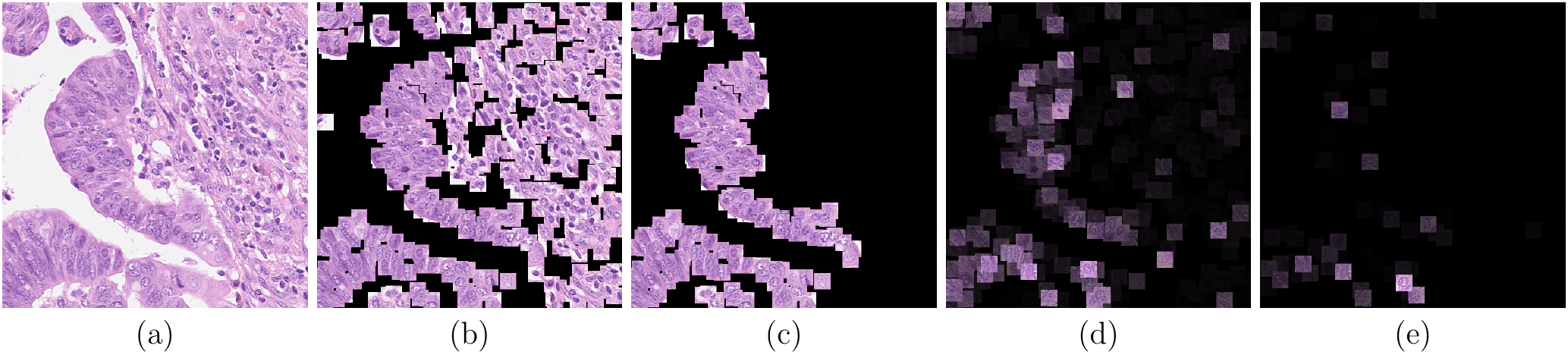}
\vskip -5pt
\caption{Colon cancer example 2: (a) H$\&$E stained histology image. (b) 27$\times$27 patches centered around all marked nuclei. (c) Ground truth: Patches that belong to the class epithelial. (d) Attention heatmap: Every patch from (b) multiplied by its attention weight. (e) Instance+max heatmap: Every patch from (b) multiplied by its score from the \textsc{instance}+$\max$ model. We rescaled the attention weights and instance scores using $a_k^\prime = (a_k - \min(\textbf{a}))/(\max(\textbf{a}) - \min(\textbf{a}))$.}
\label{fig:colon_cancer_heatmap_2}
\end{figure*}

\begin{figure*}[!hbpt]
\centering
\includegraphics[width=\columnwidth]{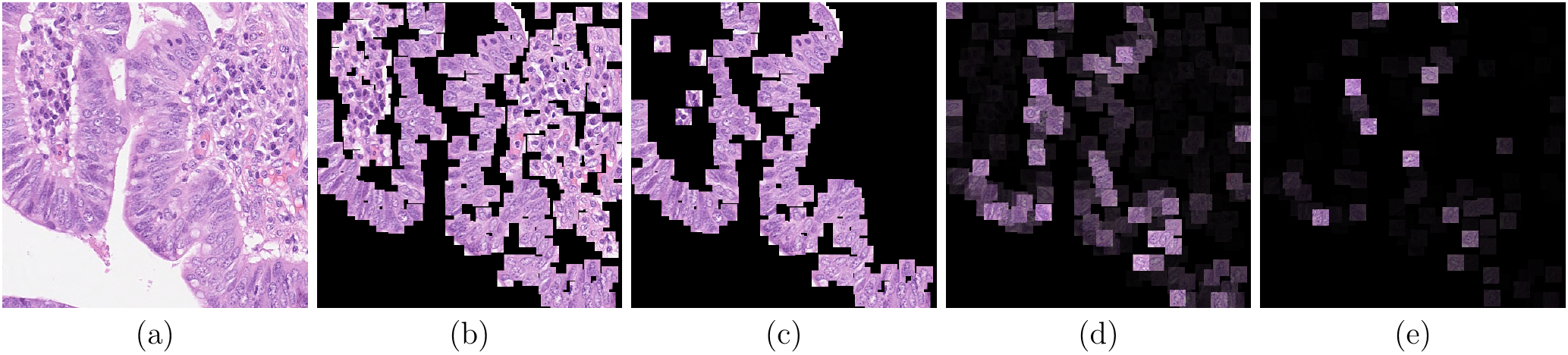}
\vskip -5pt
\caption{Colon cancer example 3: (a) H$\&$E stained histology image. (b) 27$\times$27 patches centered around all marked nuclei. (c) Ground truth: Patches that belong to the class epithelial. (d) Attention heatmap: Every patch from (b) multiplied by its attention weight. (e) Instance+max heatmap: Every patch from (b) multiplied by its score from the \textsc{instance}+$\max$ model. We rescaled the attention weights and instance scores using $a_k^\prime = (a_k - \min(\textbf{a}))/(\max(\textbf{a}) - \min(\textbf{a}))$.}
\label{fig:colon_cancer_heatmap_3}
\end{figure*}

\paragraph{Additional results} In Figures \ref{fig:colon_cancer_heatmap}, \ref{fig:colon_cancer_heatmap_2} and \ref{fig:colon_cancer_heatmap_3} five images are presented: (a) a full H\&E image, (b) all patches containing cells, (c) positive patches, (d) a heatmap given by the attention mechanism, (e) a heatmap given by the Instance+max. We rescaled the attention weights and instance scores using $a_k^\prime = (a_k - \min(\textbf{a}))/(\max(\textbf{a}) - \min(\textbf{a}))$.

\end{document}